%% file: root.tex
\documentclass[9pt, onecolumn]{article}

\usepackage[margin=1in]{geometry}  

\usepackage{svg}
\usepackage{url}
\usepackage{cite}
\usepackage{array}
\usepackage{xcolor}
\usepackage{pifont}
\usepackage{acronym}
\usepackage{leftidx}
\usepackage{csquotes}
\usepackage{colortbl}
\usepackage{textcomp}
\usepackage{graphicx}
\usepackage{textcomp}
\usepackage{stfloats}
\usepackage{verbatim}
\usepackage{listings}
\usepackage{graphicx}
\usepackage{multirow}
\usepackage{booktabs}
\usepackage{algorithm}
\usepackage{subcaption}
\usepackage{algorithmic}
\usepackage{amsmath,amsfonts}
\usepackage{newtxtext,newtxmath}
\usepackage[caption=false,font=normalsize,labelfont=sf,textfont=sf]{subfig}

\input{headers}

\begin{document}

\title{Unveiling the Potential of iMarkers: \\ Invisible Fiducial Markers for Advanced Robotics}


\author{
    Ali Tourani$^{1,3}$, Deniz Işınsu Avşar$^{2,3}$, Hriday Bavle$^{1}$, Jose Luis Sanchez-Lopez$^{1}$, \\ Jan P.F. Lagerwall$^{2}$, and Holger Voos$^{1}$
    \thanks{$^{1}$Authors are with the Automation and Robotics Research Group, Interdisciplinary Centre for Security, Reliability, and Trust (SnT), University of Luxembourg, Luxembourg. Holger Voos is also associated with the Faculty of Science, Technology, and Medicine, University of Luxembourg, Luxembourg. \tt{\small{\{ali.tourani, hriday.bavle, joseluis.sanchezlopez, holger.voos\}}@uni.lu}}
    \thanks{$^{2}$Authors are with the Department of Physics \& Materials Science, University of Luxembourg, Luxembourg. {\tt\small\{{deniz.avsar, jan.lagerwall\}@uni.lu}}}
    \thanks{$^{3}$Authors are with the Institute for Advanced Studies, University of Luxembourg, Luxembourg.}
    \thanks{*This work was partially funded by the Institute of Advanced Studies (IAS) of the University of Luxembourg (project TRANSCEND), and the Fonds National de la Recherche of Luxembourg (FNR) (project C22/IS/17387634/DEUS)}
    \thanks{*For the purpose of Open Access, and in fulfilling the obligations arising from the grant agreement, the author has applied a Creative Commons Attribution 4.0 International (CC BY 4.0) license to any Author Accepted Manuscript version arising from this submission.}
}

\markboth{Journal of \LaTeX\ Class Files,~Vol.~14, No.~8, August~2021}%
{Shell \MakeLowercase{\textit{et al.}}: A Sample Article Using IEEEtran.cls for IEEE Journals}


\maketitle

\input{sections/00-abstract}


\input{sections/01-intro}
\input{sections/02-sota}
\input{sections/03-imarker}
\input{sections/03-method}
\input{sections/04-benchmark}
\input{sections/05-discussion}
\input{sections/06-conclusion}

\section*{Acknowledgments}
The authors would like to thank Marco Giberna, Dr. Tadej Emersic, Dr. Xu Ma, and Aabhash Bhandari for their assistance in conducting experiments.


\bibliographystyle{IEEEtran}
\bibliography{root}

\newpage



\end{document}

%% file: headers.tex
\acrodef{IR}{Infrared}
\acrodef{UV}{Ultraviolet}
\acrodef{MR}{Mixed Reality}
\acrodef{FoV}{Field of View}
\acrodef{NIR}{Near Infrared}
\acrodef{AR}{Augmented Reality}
\acrodef{6DoF}{6-Degrees-of-Freedom}
\acrodef{ROS}{Robot Operating System}
\acrodef{GUI}{Graphical User Interface}
\acrodef{LiDAR}{Light Detection And Ranging}
\acrodef{ORB}{Oriented FAST and Rotated BRIEF}
\acrodef{CSR}{Cholesteric Spherical Reflector}
\acrodef{SLAM}{Simultaneous Localization and Mapping}
\acrodef{VSLAM}{Visual Simultaneous Localization and Mapping}

\newcommand{\cmark}{\ding{51}}
\newcommand{\xmark}{\ding{55}}

\definecolor{red}{HTML}{fd8f8f}
\definecolor{greend}{HTML}{57e377}
\definecolor{greenl}{HTML}{b8fb8a}
\definecolor{yellow}{HTML}{fefdb4}
\definecolor{orange}{HTML}{ffd5ab}

\colorlet{red}{red!50}
\colorlet{yellow}{yellow!50}
\colorlet{greenl}{greenl!50}
\colorlet{greend}{greend!50}
\colorlet{orange}{orange!50}

\lstdefinelanguage{yaml}{
  float=tp,
  keywords={true,false,null,y,n},
  keywordstyle=\color{blue}\bfseries,
  basicstyle=\ttfamily\small,
  comment=[l]{\%},
  morecomment=[s]{/*}{*/},
  commentstyle=\color{gray}\itshape\footnotesize,
  stringstyle=\color{orange},
  showstringspaces=false,
  alsoletter={-}
}

%% file: sections/00-abstract.tex
\begin{abstract}
Fiducial markers are widely used in robotics for navigation, object recognition, and scene understanding.
While offering significant advantages for robots and Augmented Reality (AR) applications, they often disrupt the visual aesthetics of environments, as they are visible to humans, making them unsuitable for many everyday use cases.
To address this gap, this paper presents ``\textit{iMarkers},'' innovative, unobtrusive fiducial markers detectable exclusively by robots and AR devices equipped with adequate sensors and detection algorithms.
These markers offer high flexibility in production, allowing customization of their visibility range and encoding algorithms to suit various demands.
The paper also introduces the hardware designs and open-sourced software algorithms developed for detecting iMarkers, highlighting their adaptability and robustness in the detection and recognition stages.
Numerous evaluations have demonstrated the effectiveness of iMarkers relative to conventional (printed) and blended fiducial markers and have confirmed their applicability across diverse robotics scenarios.
\end{abstract}

%% file: sections/01-intro.tex
\section{Introduction}
\label{sec_intro}

Fiducial markers, \textit{i.e.,} artificial landmarks with distinguishable graphical patterns, are widely used in robotics and computer vision to offer reliable object identification, localization, and pose estimation \cite{markersurvey}.
As markers provide better-defined reference points than naturally available ones, they are ``\textit{simple-yet-effective}'' solutions when consistent feature matching and precise pose retrieval are essential.
Additionally, augmenting information with markers and employing appropriate systems to extract it can benefit various domains, including \ac{AR} and robotics \cite{costa2024assessment}.
For instance, markers can narrow the range of objects a robot needs to recognize and provide real-time pose information for tracking and localization.
Thus, decoding data from markers detected using the robot's camera is less costly than processing the entire visual scene that the robot can see.

\begin{figure}[t]
  \centering
  \includegraphics[width=.9\columnwidth]{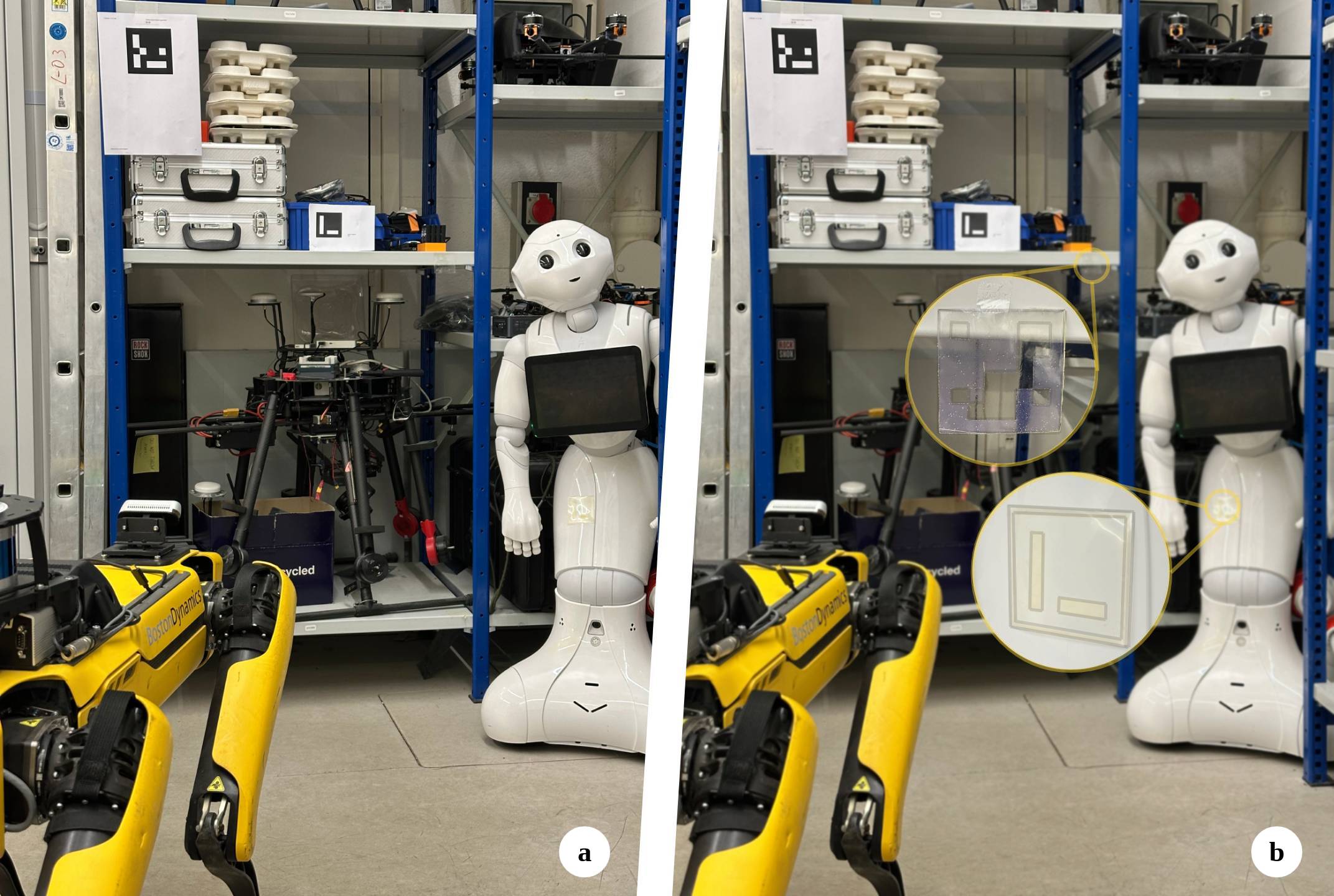}
  \caption{A legged robot observing an environment labeled with multiple \textbf{a)} printed fiducial markers and \textbf{b)} the proposed iMarkers, highlighted and magnified.}
  \label{fig_overall}
\end{figure}

Although they can convey various advantages, seeing multiple objects labeled with printed markers can be aesthetically undesirable and destructive.
In other words, many would prefer to avoid seeing printed tags around them and want the robots to operate effectively without spoiling the zones with these artificial landmarks.
Evidence also shows that visible markers can disrupt natural gaze behavior \cite{ayala2024does}.
Consequently, and building upon the idea introduced in \cite{schwartz2021linking} for using \acp{CSR} to create imperceptible landmarks, the authors explored the use of team-developed CSRs and propose a new generation of robotics fiducial markers, dubbed \textit{iMarker}\footnote{\url{https://snt-arg.github.io/iMarkers/}}.
This is achieved by replacing the printing-ink pigment used in common markers with spherical-shell \acp{CSR}.
Due to the peculiar behavior of \acp{CSR}, iMarkers can be indistinguishable to human eyes but observable by robots if a dedicated sensor is provided.
Moreover, such markers can be fabricated to be detected only within specific wavelength ranges to avoid visual clutter.
Fig.~\ref{fig_overall} shows that, in qualitative comparisons, iMarkers are hardly noticeable to the naked eye or in standard camera frames, compared with traditional printed fiducial markers.

The motivation of this paper is to demonstrate the role of iMarkers in robotics by presenting their design and detection procedures and benchmarking their applicability in real-world robotic scenarios.
Moreover, the work aims to validate the value of iMarkers in simplifying challenging robotic perception tasks, such as detecting transparent and reflective surfaces (\textit{e.g.,} glass, windows, and mirrors).
Hence, the leading contributions of the paper are summarized below:

\begin{itemize}
    \item Introducing iMarkers for robotics and \ac{AR} applications; invisible fiducial markers highly flexible in design and undetectable by the naked eye,
    \item Proposing multiple sensor solutions and open-source software algorithms for detection and 6DoF pose estimation of iMarkers of various designs; and
    \item Assessing their applicability relative to traditional fiducial markers, with emphasis on scenarios where conventional printed markers are ineffective.
\end{itemize}

%% file: sections/02-sota.tex
\section{Literature Review}
\label{sec_related_markers}

\input{sections/sota/traditional}
\input{sections/sota/blended}
\input{sections/sota/gap}

%% file: sections/sota/traditional.tex
\subsection{Ordinary Fiducial Markers}
\label{sec_related_normal}

Fiducial marker libraries are expected to enable rapid detection, reliable recognition, and efficient information decoding.
Ordinary markers, mainly printed on paper, can be classified into \textit{non-square} (designed as circles, point sets, or arbitrary visual patterns), \textit{square} (or matrix-based), and \textit{hybrid} variations.
In non-square markers, identification and data decoding generally occur \textit{w.r.t.} the center of their geometric shape.
InterSense \cite{intersense} and WhyCode \cite{whycode} are circular-shaped libraries in which the correspondence point detection is the tricky part.
Dot-shaped libraries, \textit{e.g.,} RUNE-Tag \cite{runetag}, provide fast detection but perform poorly in noisy conditions.
Other arbitrary, shape-free markers are primarily used in \ac{6DoF} pose estimation.
Despite robust detection, they face challenges, including inter-marker-encoded dot confusion and steep viewing-angle recognition.

Square fiducial markers, on the other hand, employ binary codes encoded in matrices and deliver four principal corner points for precise pose estimation.
They require lower computational cost for recognition, enabling faster detection algorithms.
ARToolkit \cite{artoolkit} is a matrix-based marker library that uses template-matching for square marker pattern determination.
However, it exhibits high reprojection errors under rotation or camera-angle changes.
AprilTag \cite{apriltag} uses gradient-based edge detection and reliable line segment extraction to identify potential tag borders, ensuring accurate marker recognition, even under challenging viewing angles.
Its geometrically driven detection pipeline, combined with built-in error-correction mechanisms, significantly reduces false positives caused by background textures and perspective distortions, making AprilTag more robust to rotation, occlusion, and warping than traditional systems.
Similarly, Garrido-Jurad \textit{et al.} introduced ArUco Marker \cite{aruco}, which supports configurable marker dictionary generation with maximized inter-marker Hamming distances.
It includes automatic marker detection with error correction via dictionary search, as well as a novel occlusion-handling method using color-based masks for robust pose estimation.
The ArUco pipeline uses adaptive thresholding to segment high-contrast regions and separate marker candidates, followed by square-shape isolation and perspective transform computation, which extract bit patterns for predefined dictionary matching.

Despite their widespread use and robust performance, these markers, whether non-square or square, remain visually intrusive and can disrupt the natural aesthetics of environments.
This limitation has motivated the development of markers that blend more seamlessly into their surroundings.

%% file: sections/sota/blended.tex
\subsection{Blended and Unobtrusive Fiducial Markers}
\label{sec_related_blended}

In contrast to paper-based markers, some works propose solutions for scenarios that require specialized detection sensors.
In this regard, ArTuga \cite{artuga} is a multimodal marker for detection in challenging environments through photometric and radiometric sensor fusion.
Seedmarker \cite{seedmarkers} considers the aesthetic properties of markers by embedding them in physical objects, \textit{e.g.,} laser-cut plates and 3D-printed tangibles.
ACMarker \cite{acmarker} is another approach that can be detected by custom sensors in underwater robotics applications.

Other approaches focus on integrating less obtrusive markers into environments to minimize their visual impact while maintaining robust detectability.
Works in \cite{schwartz2021linking, agha2022unclonable} introduced the potential of creating imperceptible visual landmarks by exploiting the optical properties of particular materials.
However, they remain primarily focused on materials and optics and do not investigate the applicability of landmarks, such as fiducial markers, for perception robustness, validity, or accuracy.
Despite briefly exploring the potential integration of such landmarks in a robotics context in \cite{vsgraphs1}, it does not provide a concrete sensing-modality design, a description of detection algorithms, or a systematic experimental evaluation.
TRDM \cite{trdm} uses randomly distributed printed dots on transparent sheets to achieve low visual saliency.
Nevertheless, geometric point matching and marker detection are tricky while integrating pairwise relationships among the detected dots.
InfraredTags \cite{infraredtags} are discreet, durable markers printed and embedded in various objects, providing invisibility and enabling rapid scanning.
However, their fabrication is limited to the \ac{IR} range, reducing design flexibility, and their thin, embedded form makes them difficult to locate visually, complicating alignment of \ac{IR} illumination and sensing during deployment.
BrightMarkers \cite{brightmarker} are 3D-printed fluorescent filaments embedded in objects and require \ac{NIR} cameras for detection.
Aircode \cite{aircode} embeds markers directly into the fabrication of objects, enabling seamless integration for use cases such as robotic grasping.
Nevertheless, it suffers from a slow decoding time (tens of seconds), which depends on the camera's viewing angle.

%% file: sections/sota/gap.tex
\subsection{Gap Identification}
\label{sec_gap}

Although the surveyed blended and unobtrusive methodologies have demonstrated effectiveness in robotics and \ac{AR} tasks by keeping markers discreet, their applicability in real-world scenarios is often constrained by the trade-off between ``\textit{invisibility}'' and ``\textit{detection speed.}''
In other words, solutions like AirCode achieve high levels of invisibility but suffer from significantly slow detection speeds.
Other crucial factors include simplicity, cost-effectiveness, versatility in detection, robust performance, and flexibility in fabrication across various light spectra.
These dimensions significantly influence the adoption of such marker solutions compared with straightforward, easily printable traditional markers.

%% file: sections/03-imarker.tex
\section{iMarkers: Concept and Comparison}
\label{sec_marker}

iMarkers offer an innovative, versatile solution to address the gaps mentioned, ensuring transparency, flexibility, and unobtrusiveness.
Given appropriate sensors and detection algorithms, they can be fabricated to achieve high contrast for detection while remaining undetectable to the naked eye.
Table~\ref{tbl_markers} highlights the characteristics of iMarkers compared to other blended methodologies.
Accordingly, although all solutions aim to remain hidden from human perception, iMarkers cover distinct aspects that set them apart:

\input{tables/tbl_markers}

\begin{figure}[!b]
    \centering
    \includegraphics[width=.8\columnwidth]{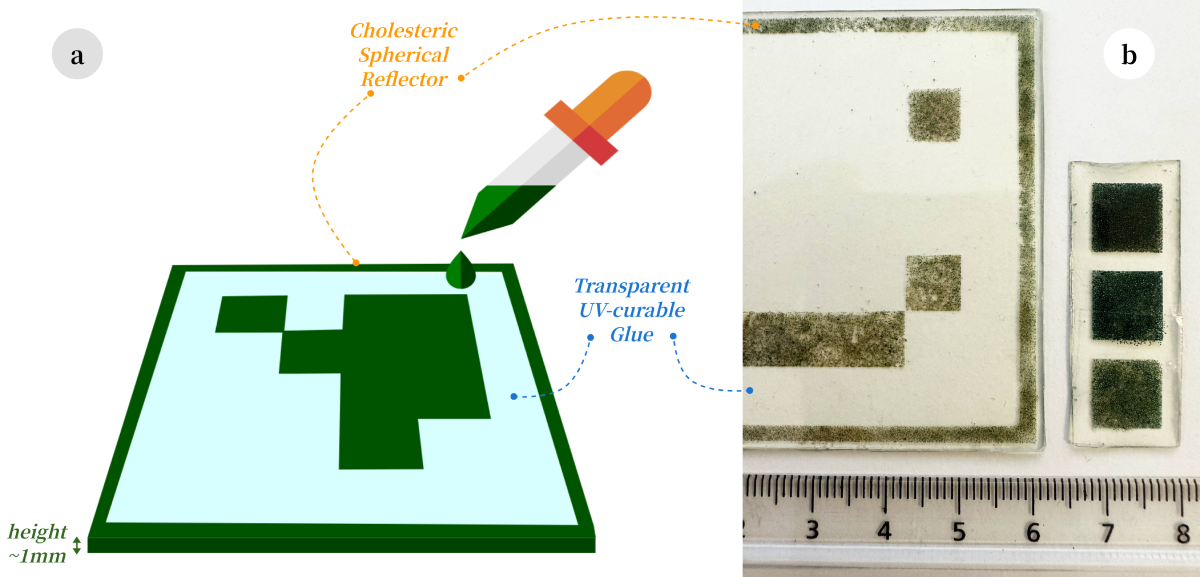}
    \caption{The role of CSRs in the design of iMarkers: \textbf{(a)} an iMarker featuring ArUco patterns filled with CSRs, \textbf{(b)} an iMarker displayed next to visible-range CSR particles, illustrating various shades of green for fabrication.}
    \label{fig_marker_material}
\end{figure}

\textbf{Core Material:}
The material used to make iMarkers is \acp{CSR} \cite{agha2022unclonable}, microscopic droplets of Cholesteric Liquid Crystals (CLCs) that self-organize into a helical structure, making them highly effective at reflecting specific wavelengths of light, including \ac{IR}, \ac{UV}, or visible.
Their spherical shape makes them omnidirectional, selective retroreflectors, ensuring detectability regardless of viewing angle.
They produce vivid, circularly polarized iridescent colors through selective light reflection, enabling clear separation between CSR and non-CSR regions.
As shown in Fig.~\ref{fig_marker_material}, iMarkers combine \acp{CSR} for the marker patterns with transparent UV-curable glue to maintain the structure of the fiducial marker.
An alternative fabrication method involves directly spraying \ac{CSR} shells onto surfaces, a promising approach currently under investigation.

\begin{figure}[!h]
    \centering
    \includegraphics[width=.8\columnwidth]{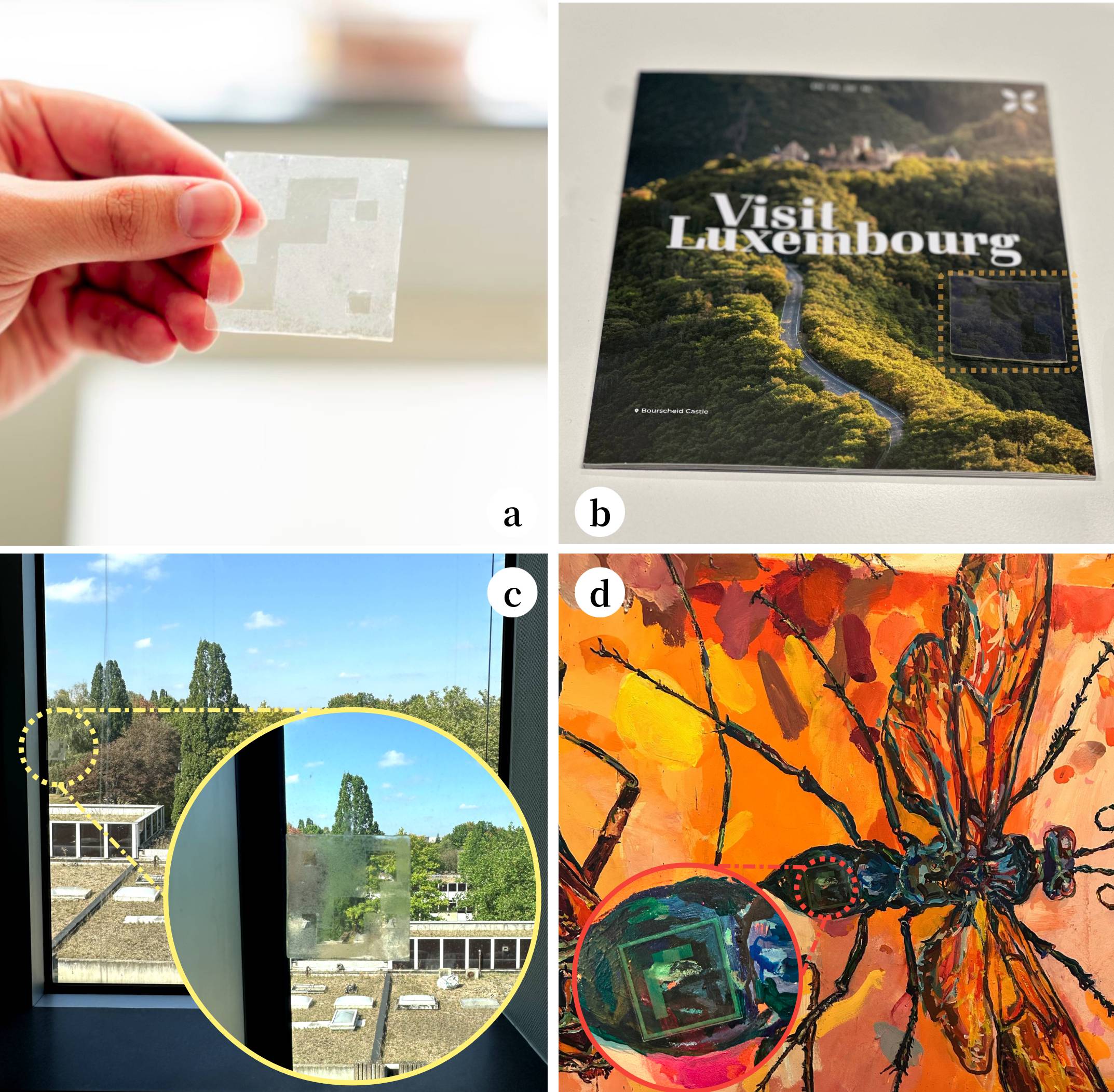}
    \caption{Versatility in iMarker production: \textbf{(a)} a $6 \times 6 ~\mathrm{cm^2}$ UV-range iMarker, \textbf{(b)} an IR-range iMarker with the same size, \textbf{(c)} the UV-range iMarker placed on a transparent surface, \textbf{(d)} a $7 \times 7 ~\mathrm{cm^2}$ dark green visible-range iMarker camouflaged with its background.}
    \label{fig_marker_variants}
\end{figure}

\textbf{Pattern Coding and Recognition:}
iMarkers support various coding patterns for recognition and information retrieval.
This paper focuses on ArUco-based \cite{aruco} markers with variant dictionaries, but other standard fiducial marker libraries can also be adapted to the iMarker design.

\textbf{Fabrication Cost and Efficiency:}
The cost and simplicity of fabrication significantly impact the widespread adoption of markers.
While AirCode requires partitioning and InfraStructs depend on a complex, layered assembly, iMarkers, as shown in Fig.~\ref{fig_marker_material}, can be fabricated using cost-effective methods such as dispensing or spray coating.

\textbf{Production Versatility:}
In contrast with other solutions, iMarkers offer production flexibility, enabling design and detection across the entire light spectrum.
As shown in Fig.~\ref{fig_marker_variants}, UV-range iMarkers remain entirely transparent and invisible to the naked eye, even when placed on a transparent surface such as glass.
IR-range iMarkers are similarly discreet, though slight scattering or subtle color effects may reveal their presence (to some extent).
Even in the visible range, iMarkers can be designed to blend with patterned backgrounds or camouflage against similarly colored surfaces, ensuring minimal visual intrusion.

\textbf{Sensor Affordability:}
Detection requirements vary across unobtrusive marker systems: AirCode relies on projector-camera setups, InfraStructs require costly Terahertz scanners, and InfraredTag needs specialized imaging modules with microprocessors.
In contrast, iMarkers can be detected using affordable sensors with standard optical components, provided the camera is compatible with the polarization and wavelength characteristics of their \ac{CSR} material.

\textbf{Detection Algorithm Performance:}
Since the sensors are designed to generate binary images, fiducial marker patterns are detected within milliseconds, ensuring real-time performance for various robotics and AR use cases.

%% file: tables/tbl_markers.tex
\begin{table*}[!t]
    \centering
    \caption{A comparison of the primary features and capabilities of various unobtrusive fiducial marker solutions.}
    \begin{tabular}{l|cccc}
        \toprule
            & \textbf{AirCode} \cite{aircode} & \textbf{InfraStructs} \cite{infrastructs} & \textbf{InfraredTags} \cite{infraredtags} & \textbf{iMarker} (ours) \\
        \midrule
            \rowcolor[gray]{0.92} \textit{Unobtrusiveness to human} & \cmark & \cmark & \cmark & \cmark \\
            \textit{Core material} & air pockets & polystyrene & IR filament & CSRs \\
            \rowcolor[gray]{0.92} \textit{Versatile pattern coding} & \xmark & \xmark & \cmark & \cmark \\
            \textit{Fabrication cost} & high & normal & low & low \\
            \rowcolor[gray]{0.92} \textit{Fabrication simplicity} & \xmark & \xmark & \cmark & \cmark \\            
            \textit{Production versatility} $^{\mathrm{*}}$ & \textit{Vis}, \textit{UV} & \textit{IR} & \textit{IR} & \textit{IR}, \textit{UV}, \textit{Vis} \\
            \rowcolor[gray]{0.92} \textit{Sensor cost} & very high & high & low & low \\
            \textit{Detection time range} & \textit{second} & \textit{second} & \textit{millisecond} & \textit{millisecond} \\
        \bottomrule
        \multicolumn{5}{l}{$^{\mathrm{*}}$\textit{IR}, \textit{UV}, and \textit{Vis} refer to Infrared, Ultraviolet, and visible ranges, respectively.}
    \end{tabular}
    \label{tbl_markers}
\end{table*}

%% file: sections/03-method.tex
\section{Sensor Design and Detection Strategies}
\label{sec_method}

Three primary solutions are proposed for iMarker detection, including ``\textit{dual-vision,}'' ``\textit{dynamic single-vision,}'' and ``\textit{static single-vision,}'' further described in the following:

\input{sections/method/dual-vision}
\input{sections/method/single-vision-dyn}
\input{sections/method/single-vision-sta}

%% file: sections/method/dual-vision.tex
\subsection{Dual-vision Solution}
\label{sec_dual}

\begin{figure}[t]
     \centering
     \includegraphics[width=.8\columnwidth]{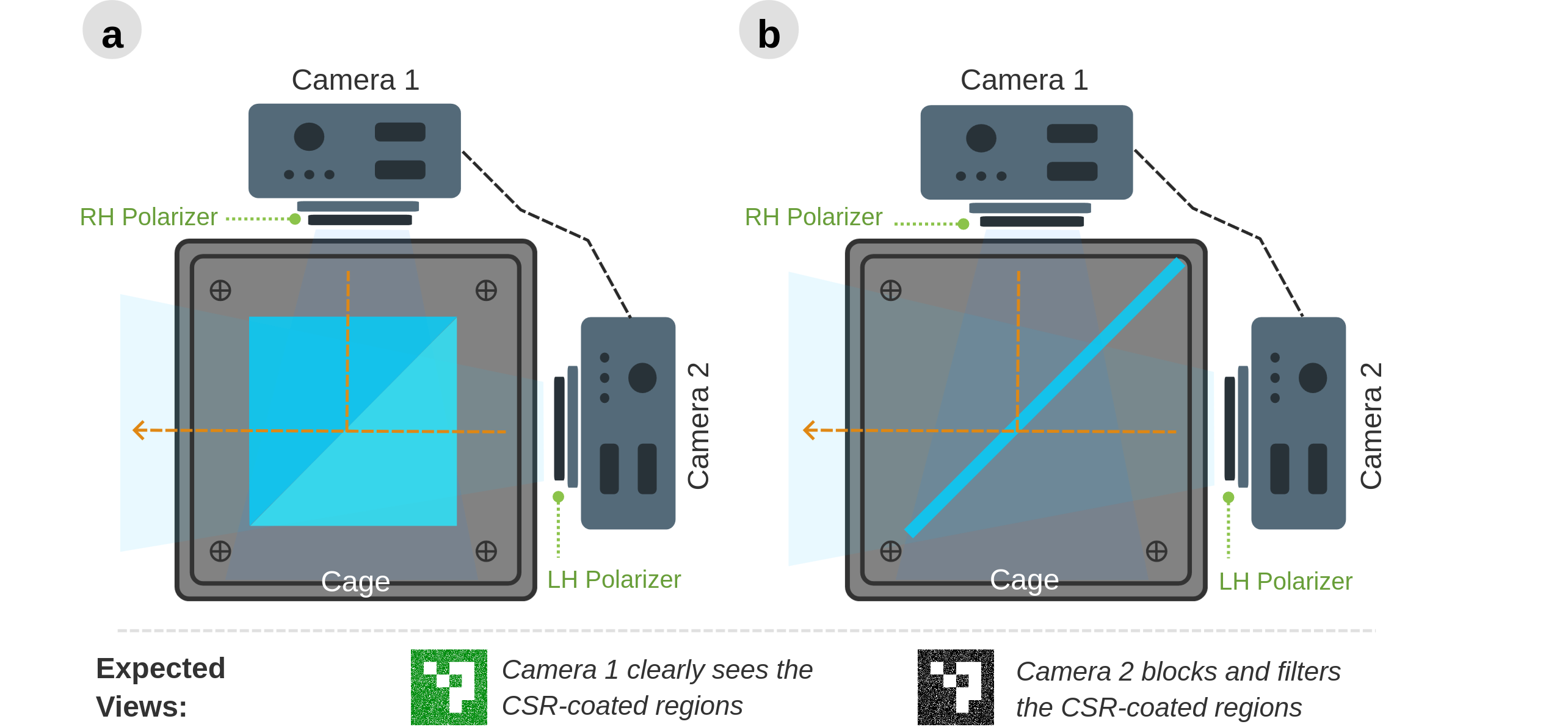}
     \caption{Dual-vision iMarker detection setups using synchronized cameras with circular polarizers and either a cube \textbf{(a)} or plate \textbf{(b)} beamsplitter.}
     \label{fig_hardware_dual}
\end{figure}

\noindent\textbf{Sensor Design.}
The first setup is a homogeneous perception system comprising two synchronized, perpendicularly positioned cameras, each facing a different surface of an optical beamsplitter.
As shown in Fig.~\ref{fig_hardware_dual}, the sensor can be assembled with either a cube or a plate beamsplitter.
The beamsplitter equally divides the incoming light (\textit{i.e.,} $50\texttt{:}50$) into two parts: the transmitted and the reflected light.
By mounting the cameras and the beamsplitter in a cage, one camera captures the scene using the transmitted light, while the other captures it using the reflected light, making two perspectives with minimal variation.
Attaching identical circular polarizers to each camera's lens effectively blocks light with the opposite circular polarization, ensuring only the desired handedness is detected.
Due to the inversion of circular polarization upon reflection at the beamsplitter, the camera capturing the reflected light detects the \enquote{opposite} circular polarization to that of the camera capturing the transmitted light.
Hence, the setup simultaneously captures the same scene with two cameras, ensuring that the \ac{CSR} regions of iMarkers in the sensor's \ac{FoV} appear blocked in one camera while remaining visible in the other (due to the opposing circular polarization of the \ac{CSR} reflections).

Note that the cameras in this setup must be synchronized to capture the same scene simultaneously (frame \(f_t^{C_1}\) captured by camera \(C_1\) at time \(t\) should depict the same scene as frame \(f_t^{C_2}\) of camera \(C_2\)).
As the cameras' frame rate and shutter speed are vital for high-speed robotic applications (\textit{e.g.,} drone navigation), integrating global-shutter cameras in this sensor is highly recommended.
The calibration procedure involves capturing a standard calibration pattern (\textit{e.g.,} a chessboard) by each camera.
The camera's intrinsic (\textit{e.g.,} focal length and distortion coefficients) and extrinsic parameters are required for iMarker pose estimation.

\input{sections/method/software/alg_dual_spatial}

\noindent\textbf{Detection Algorithm.}
Since the non-CSR regions of the scene lack circular polarization and appear identical in both cameras, simple frame subtraction can detect iMarkers with high contrast, as shown in Algorithm~\ref{alg_double_subtract}.
Accordingly, the iMarker detection procedure requires \enquote{\textit{spatial subtraction}} of the synchronized frames captured by each camera.
Having \(\mathbf{F}_{C_1}\) and \(\mathbf{F}_{C_2}\) as the synchronized frame sets captured by calibrated cameras $C_1$ and $C_2$, each frame \(f_1 \in \mathbf{F}_{C_1}\) finds its corresponding synchronized frame \(f_2 \in \mathbf{F}_{C_2}\).
The next stage is to align $f_2$ with respect to $f_1$ using the alignment parameters $h$ (captured during calibration), where $h$ is the homography matrix computed from matching \ac{ORB} features.
Subtracting the \enquote{aligned} $f_2$ from the \enquote{original} $f_1$ frame results in the final binary image $f_p$ containing the input images' dissimilar parts.
Since the \ac{CSR} region of iMarkers is visible in one frame and blocked in the other, only the patterns created by \acp{CSR} are expected to remain in $f_p$.
Then, a thresholding process with a value of $\theta$, followed by post-processing, is applied to $f_p$ to generate a binary frame that enhances the iMarker patterns' visibility.
The final processed frame contains potential iMarkers \(\mathbf{M}\) at time $t$.

%% file: sections/method/software/alg_dual_spatial.tex
\begin{algorithm}[t]
    \caption{\small{Dual-vision iMarker detection.}}
    \begin{algorithmic}
        \REQUIRE frame-sets $F_1$ and $F_2$ from cameras $C_1$ and $C_2$
        \ENSURE list of detected fiducial markers $M$
        \STATE $M \gets []$
        \STATE $p_1 \gets$ calibrate $C_1$
        \STATE $p_2 \gets$ calibrate $C_2$
        \STATE $h \gets$ align $(F_1, F_2)$ considering $p_1$ and $p_2$
        \WHILE{$f_1$ in $F_1$}
            \STATE $f_2 \gets$ corresponding sync frame in $F_2$
            \STATE $f_2 \gets$ align $f_2$ based on $f_1$ using $h$
            \STATE $f_p \gets$ $f_2 - f_1$ \COMMENT{final subtracted image}
            \STATE $f_p \gets$ threshold $f_p$
            \STATE $f_p \gets$ post-process $f_p$ \COMMENT{erosion $+$ Gaussian blur}
            \IF{marker $m$ found in $f_p$}
                \STATE add $m$ to $M$
            \ENDIF
        \ENDWHILE
        \RETURN $M$
    \end{algorithmic}
    \label{alg_double_subtract}
\end{algorithm}

%% file: sections/method/single-vision-dyn.tex
\subsection{Dynamic Single-vision Setup}
\label{sec_single-dynamic}

\begin{figure}[!b]
     \centering
     \includegraphics[width=.7\columnwidth]{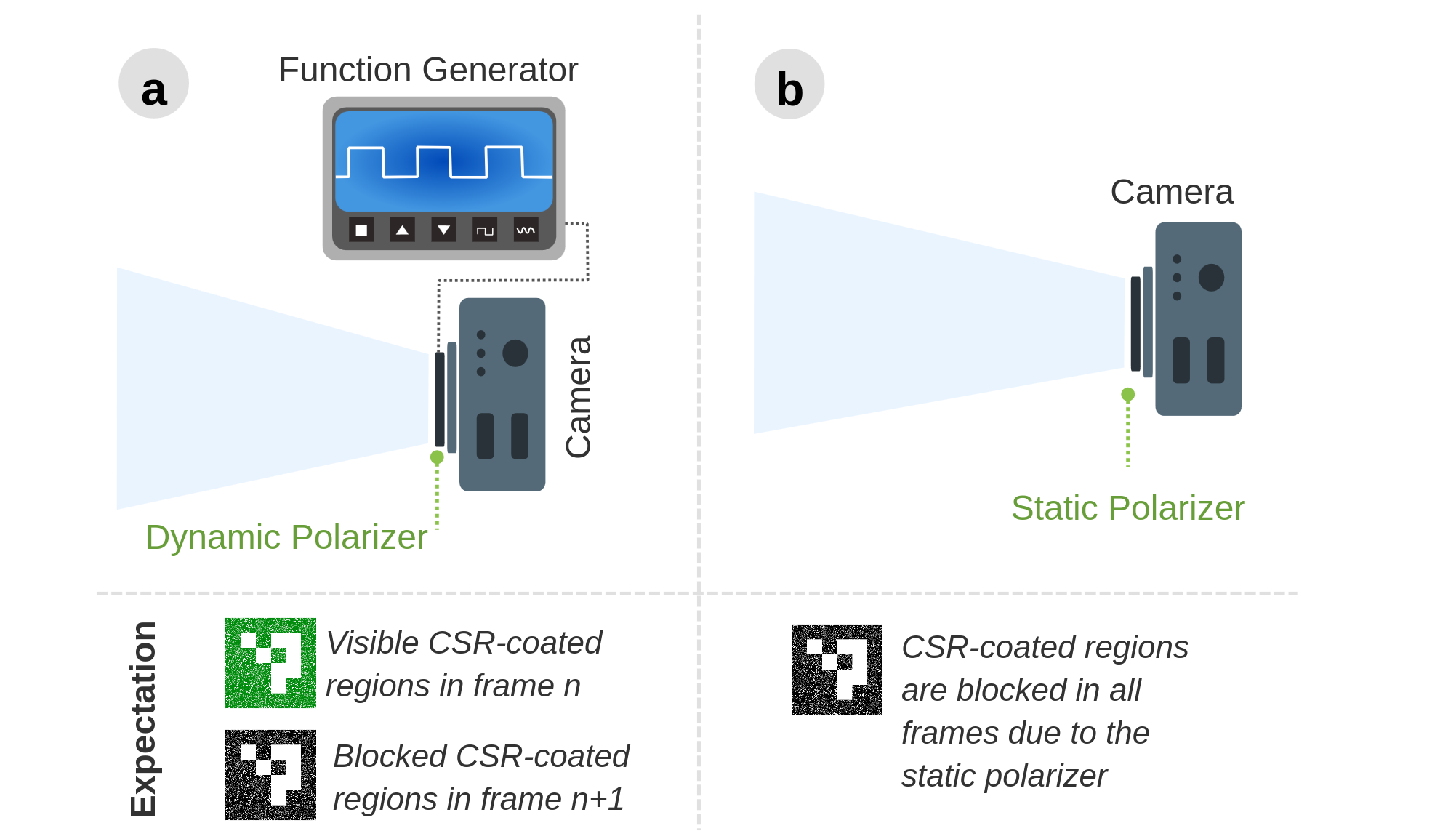}
     \caption{Single-vision setup variants designed for iMarker detection, each containing a camera with attached dynamic \textbf{(a)} or static \textbf{(b)} polarizers.}
     \label{fig_hardware_single}
\end{figure}

\noindent\textbf{Sensor Design.}
The second approach offers a single camera equipped with a switchable (dynamic) circular polarizer alternating between left- and right-handed polarization (Fig.~\ref{fig_hardware_single}a), which will be described in detail in a forthcoming paper.
In brief, it comprises a fixed circular polarizer paired with a nematic liquid crystal film that functions as a $\lambda/2$-plate in its relaxed (OFF) state and loses its optical functionality when an electric field is applied (ON).
A $\lambda/2$-plate reverses circular polarization; therefore, by switching the liquid crystal film between its ON and OFF states (controlled by a function generator synchronized with the camera's frame rate), the camera sequentially captures frames with right- and left-handed polarization.
In an ideal setup, camera frame $f_t$ captures the scene with a particular polarization (left- or right-handed), and frame $f_{t+1}$ captures it with the opposite polarization.
Thus, the function generator should send signal $\mathbf{S}_{g}$ to trigger the liquid crystal film at time $t$ as:

\begin{equation}
    \mathbf{S_{g}} = 
    \begin{cases} 
        \text{on,} & \text{for } t = 1, 3, \ldots, 2n-1, \\
        \text{off,} & \text{for } t = 2, 4, \ldots, 2n.
    \end{cases}
    \label{eq_cpolarizer}
\end{equation}

The liquid crystal cell's OFF-to-ON switching time is negligible, but the ON-to-OFF relaxation is slower.
Thus, intermediate frames with undefined polarization at high frame rates may need to be discarded during iMarker detection.
Once optimized for the liquid crystal cell, it ensures that \ac{CSR}-coated regions in the iMarkers are blocked in one processed frame and visible in the next.

\noindent\textbf{Detection Algorithm.}
Inspired by the \enquote{\textit{spatial}} frame processing in the dual-vision setup, this approach employs a frame-level algorithm that uses \enquote{\textit{temporal}} subtraction to detect the iMarker patterns.
Assuming the liquid crystal film operates sufficiently fast \textit{w.r.t} the video frame rate to produce sequential frames with alternating polarization states (Equation~\ref{eq_cpolarizer}), the consecutive (temporal) frame subtraction algorithm to detect iMarkers using a switchable polarizer is shown in Algorithm~\ref{alg_single_subtract}.
Here, each frame $f_{t-1}^\text{~off}$ and its subsequent frame $f_t^\text{~on}$, captured with alternating polarization states, are subtracted to produce a grayscale image $f_p$, which can then be enhanced through post-processing to facilitate iMarker detection.
Additionally, the current frame is stored for subtraction from its succeeding frame in the next iteration.
As with the previous approaches, the final stage involves detecting markers $\mathbf{M}$ within the frame $f_p$.

\input{sections/method/software/alg_single_temporal}

%% file: sections/method/software/alg_single_temporal.tex
\begin{algorithm}[t]
    \caption{\small{Dynamic single-vision iMarker detection.}}
    \label{alg_single_subtract}
    \begin{algorithmic}
        \REQUIRE frame-set $F$ from the camera
        \ENSURE list of detected fiducial markers $M$
        \STATE $M \gets []$
        \STATE $f_{prev} \gets null$
        \WHILE{$f_t$ in $F$}
            \STATE $f_p \gets$ $f_t - f_{prev}$ \COMMENT{final subtracted image}
            \STATE $f_p \gets$ post-process $f_p$ \COMMENT{erosion $+$ Gaussian blur}
            \STATE $f_{prev} \gets f_t$
            \IF{marker $m$ found in $f_p$}
                \STATE add $m$ to $M$
            \ENDIF
        \ENDWHILE
        \RETURN $M$
    \end{algorithmic}
\end{algorithm}

%% file: sections/method/single-vision-sta.tex
\subsection{Static Single-vision Setup}
\label{sec_single}

\noindent\textbf{Sensor Design.}
Another configuration offers a solution for iMarker detection using a single camera equipped with a static circular polarizer attached to its lens (Fig.~\ref{fig_hardware_single}b).
The polarizer is selected to block the circular polarization reflected by the \acp{CSR} forming the iMarker pattern.
For instance, a left-handed circular polarizer will block iMarker regions with \acp{CSR} reflecting right-handed polarization, and vice versa.
Thus, a visible-range iMarker with \acp{CSR} reflecting right-hand polarized light on a same-color background (\textit{e.g.,} green), indistinguishable to the human eye or standard cameras, becomes detectable as the sensor's left-handed polarizer blocks the \ac{CSR} regions, revealing the camouflaged iMarker patterns.
This design is well suited to detecting visible-range iMarkers camouflaged in the environment, and it can also be used for \ac{IR}/\ac{UV}-range iMarkers with appropriate cameras.

\noindent\textbf{Detection Algorithm.}
Unlike the dual-vision solution, the algorithm for the static single-vision setup does not eliminate the background.
However, it simplifies the integration of an iMarker detection sensor by shifting more computation to the detector, achieving a practical balance between complexity and performance.
For the visible-range camouflaged iMarkers, the procedure is shown in Algorithm~\ref{alg_single_mask}.
Accordingly, detecting iMarkers \(\mathbf{M}\) in a frame \(f \in \mathbf{F}\) involves \textit{RGB} to \textit{HSV} color space conversion to isolate the color range corresponding to the marker.  
The resulting mask frame \(f_p\) highlights the target color range \(c_r\) as black pixels, with all other areas appearing as white.
With the assistance of the polarizer, the mask frame \(f_p\) highlights the blocked \ac{CSR} regions of an iMarker with color \(c\) (\textit{e.g.,} green) positioned against a background of the same color.
Subsequent stages apply post-processing to eliminate noise, followed by detecting marker patterns \(\mathbf{M}\) within \(f_p\).
Detecting \ac{UV}/\ac{IR}-range iMarkers follows the same stages as the visible-range detection, provided that a \ac{UV}/\ac{IR}-range camera is used.
However, \enquote{\textit{color masking}} is unnecessary because the input is grayscale, and \enquote{\textit{range thresholding}} is used to identify intensity variations in \ac{CSR} regions, thereby highlighting the iMarker patterns, as introduced in Algorithm~\ref{alg_single_thresh}.

\input{sections/method/software/alg_single_mask}
\input{sections/method/software/alg_single_mask_uv}

%% file: sections/method/software/alg_single_mask.tex
\begin{algorithm}[t]
    \caption{\small{Static single-vision iMarker detection (masking).}}
    \label{alg_single_mask}
    \begin{algorithmic}
        \REQUIRE frame-set $F$ from the camera ($RGB$)
        \ENSURE list of detected fiducial markers $M$
        \STATE $M \gets []$
        \STATE $r \gets (low,high)$ \COMMENT{demanded color range in $HSV$}
        \WHILE{$f$ in $F$}
            \STATE $f_p \gets$ convert $f$ to color space $HSV$
            \STATE $f_p \gets$ filter $f_p$ with the range $r$
            \STATE $f_p \gets$ post-process $f_p$ \COMMENT{erosion $+$ Gaussian blur}
            \IF{marker $m$ found in $f_p$}
                \STATE add $m$ to $M$
            \ENDIF
        \ENDWHILE
        \RETURN $M$
    \end{algorithmic}
\end{algorithm}

%% file: sections/method/software/alg_single_mask_uv.tex
\begin{algorithm}[t]
    \caption{\small{Static single-vision iMarker detection (thresholding).}}
    \label{alg_single_thresh}
    \begin{algorithmic}
        \REQUIRE frame-set $F$ from the camera ($grayscale$)
        \ENSURE list of detected fiducial markers $M$
        \STATE $M \gets []$
        \STATE $r \gets (low,high)$ \COMMENT{demanded gray-level range in $grayscale$}
        \WHILE{$f$ in $F$}
            \STATE $f_p \gets$ filter $f_p$ with the range $r$
            \STATE $f_p \gets$ post-process $f_p$ \COMMENT{erosion $+$ Gaussian blur}
            \IF{marker $m$ found in $f_p$}
                \STATE add $m$ to $M$
            \ENDIF
        \ENDWHILE
        \RETURN $M$
    \end{algorithmic}
\end{algorithm}

%% file: sections/04-benchmark.tex
\section{Benchmarking and Evaluation}
\label{sec_benchmark}

\input{sections/evaluation/implement}
\input{sections/evaluation/invi_det}
\input{sections/evaluation/detection}
\input{sections/evaluation/low-light}
\input{sections/evaluation/recog_speed}

%% file: sections/evaluation/implement.tex
\subsection{Operational Implementation}
\label{sec_implement}

\begin{figure}[!t]
     \centering
     \includegraphics[width=.8\columnwidth]{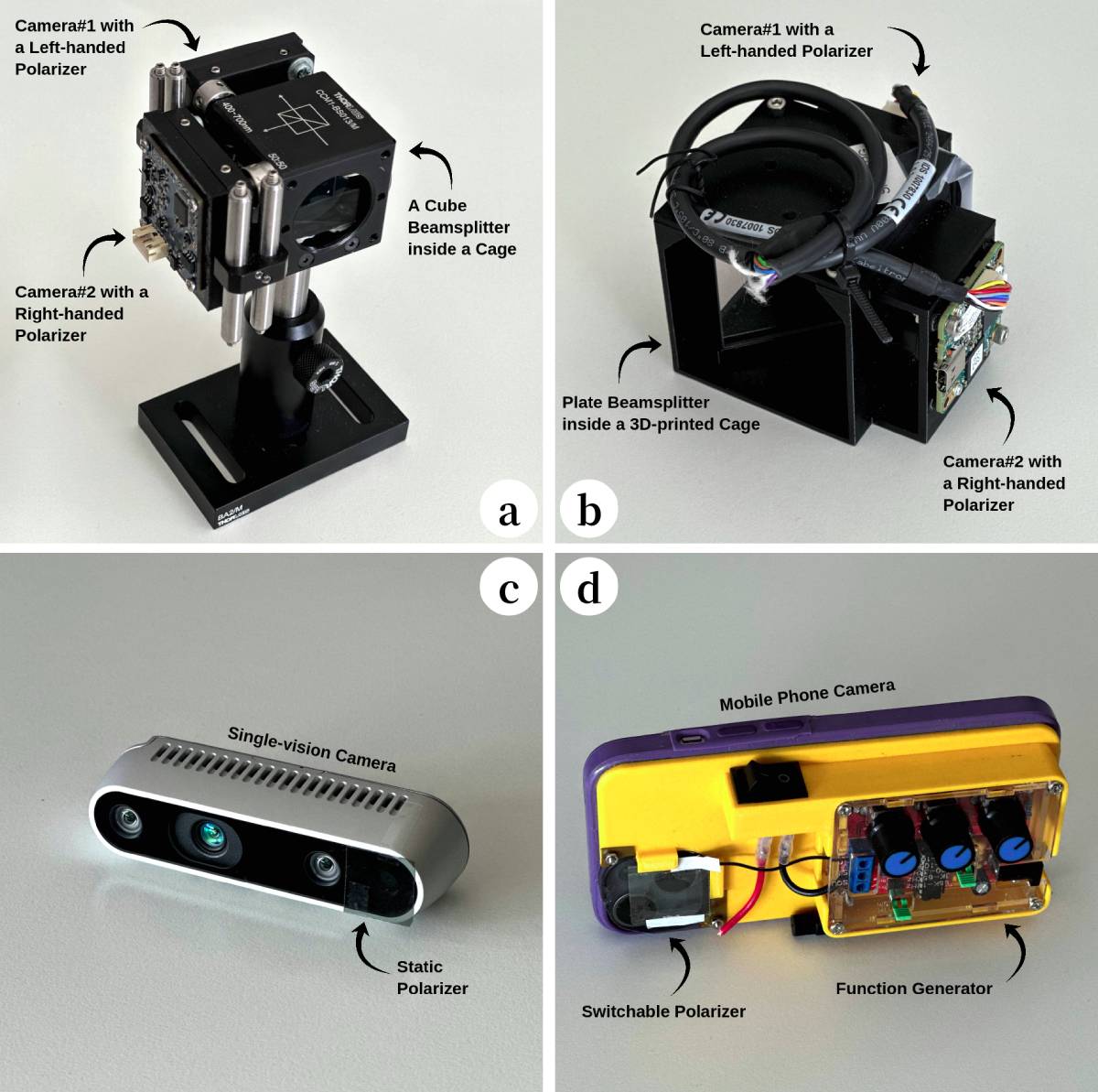}
     \caption{Variant iMarker detection sensors, including the dual-vision setup with cube \textbf{(a)} and plate \textbf{(b)} beamsplitters, and single-vision setups with static \textbf{(c)} and dynamic \textbf{(d)} polarizers.}
     \label{fig_benchmark_hardware}
\end{figure}

\noindent\textbf{Sensor.}
Fig.~\ref{fig_benchmark_hardware} depicts the different sensor configurations developed for iMarker detection.
The dual-vision setup shown in Fig.~\ref{fig_benchmark_hardware}a comprises two ELP 8MP HD camera modules with $75^\circ$ \ac{FoV} facing a Thorlabs polarizing cube beamsplitter with a 50:50 split ratio, all mounted in a compact cage assembly.
The other dual-vision setup in Fig.~\ref{fig_benchmark_hardware}b comprises two synchronized U3-3271LE iDS cameras, aligned with a Thorlabs polarizing plate beamsplitter, housed within a custom-designed 3D-printed cage.
The static single-vision setup in Fig.~\ref{fig_benchmark_hardware}c uses an Intel® RealSense™ D435 camera with a left-handed circular polarizer affixed to its RGB lens, effectively blocking the right-handed \ac{CSR}-coated areas of camouflaged iMarkers.
Finally, the dynamic single-vision sensor setup in Fig.~\ref{fig_benchmark_hardware}d incorporates a switchable polarizer controlled by a function generator mounted on an iPhone 13 case and aligned with its camera.
While this prototype relies on manual polarizer switching, we are investigating a version in which the detection software dynamically synchronizes the camera and the function generator to adjust the polarization.

\noindent\textbf{Software.}
To support the detection algorithms outlined in Section~\ref{sec_method}, an open-sourced\footnote{\url{https://snt-arg.github.io/iMarkers/}} iMarker detection software framework has been developed.
The framework is implemented in Python and features a $\mathrm{ROS2}$ interface, enabling straightforward integration into robotic applications.

%% file: sections/evaluation/invi_det.tex
\subsection{Qualitative Evaluation}
\label{sec_challenging}

Given the complexity of measuring aesthetic perception, this section aims to \textbf{visually assess} iMarkers and showcase their detection performance.
To achieve this, various iMarkers were strategically placed in fixed real-world locations and captured by multiple sensors, highlighting their seamless integration across diverse settings.
While aesthetic perception is inherently subjective, the visual comparisons in Fig.~\ref{fig_overall} and Fig.~\ref{fig_marker_variants} provide practical, real-world evidence supporting the unobtrusiveness of iMarkers across diverse environments.

\input{figures/detection}

Fig.~\ref{fig_benchmark} depicts several qualitative results, showing the versatility of iMarkers across different spectral ranges and detection setups.
Accordingly, Fig.~\ref{fig_detection_sv_st} shows a green iMarker in the visible range discreetly embedded on a similarly colored background, demonstrating its visual camouflage and unobtrusiveness.
In this scenario, the static single-vision setup was employed to capture the scene, with the setup's polarizer effectively blocking the CSR regions that form the inner iMarker pattern, thereby enhancing contrast and improving color separation for detection.
The detection and pose estimation outcome was obtained using Algorithm~\ref{alg_single_mask}.
Fig.~\ref{fig_detection_sv_dy} shows the output of the dynamic single-vision setup applied to a visible-range iMarker under indoor conditions with controlled illumination.
In contrast, Fig.~\ref{fig_detection_sv_dy_o} presents the corresponding outdoor results, where varying sunlight, shadows, reflections, and background clutter introduce additional challenges for reliable iMarker detection.
Notably, the system properly isolates the marker region with high contrast, enabling reliable pattern recognition and data decoding.
In both dynamic single-vision scenarios, detection is achieved through temporal subtraction between frames $f_{t-1}^\text{~off}$ and $f_t^\text{~on}$ following the procedure described in Algorithm~\ref{alg_single_subtract}.
Fig.~\ref{fig_detection_dv_pbs} presents the output of the dual-vision setup applied to a visible-range iMarker placed on a patterned surface under low-illumination conditions.
In this configuration, both cameras simultaneously capture the same scene, with the left camera acquiring frame $f_l$ through a left-handed polarizer and the right camera capturing frame $f_r$ through a right-handed polarizer.
Marker detection was performed using Algorithm~\ref{alg_double_subtract}, which applies spatial subtraction between the polarized image pairs.
The system effectively suppresses background texture and illumination artifacts, preserving high contrast and reliable detection under challenging lighting conditions.
As illustrated in Fig.~\ref{fig_detection_sv_uv}, the UV-range iMarker placed on a transparent surface is virtually invisible to the naked eye.
To reveal the inner patterns of the iMarker, near-UV illumination was applied, and the same scene was captured with a UV-range camera, as shown in the middle sub-figure.
The method introduced in Algorithm~\ref{alg_single_thresh} facilitates marker detection by applying a tailored thresholding strategy to the UV image, enabling reliable extraction of iMarker's invisible patterns.

Across the evaluated scenarios, iMarkers consistently remain detectable under various environmental and sensing conditions, including indoor/outdoor settings, varying illumination, patterned backgrounds, and transparent surfaces.
Despite reduced visual contrast, the proposed sensing and detection strategies reliably isolate the marker regions and preserve sufficient pattern contrast for information decoding.

%% file: figures/detection.tex
\begin{figure}[h!]
     \centering
     \begin{subfigure}{0.6\columnwidth}
        \centering
        \begin{subfigure}{0.32\columnwidth}
            \includegraphics[width=\linewidth]{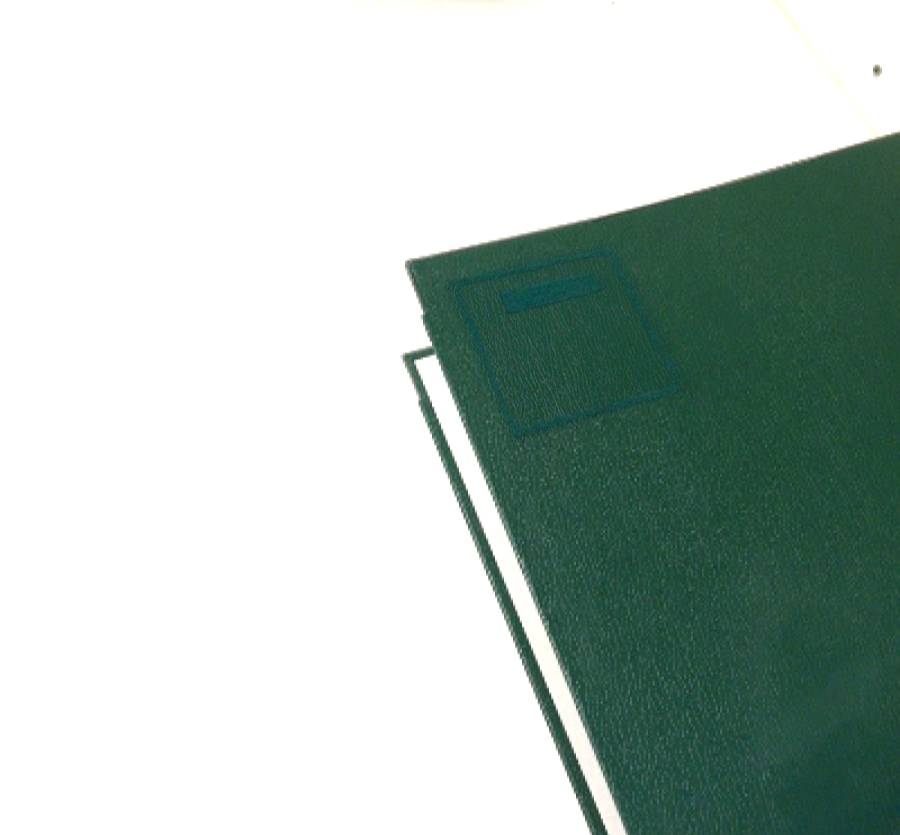}
        \end{subfigure}
        \begin{subfigure}{0.32\columnwidth}
            \includegraphics[width=\linewidth]{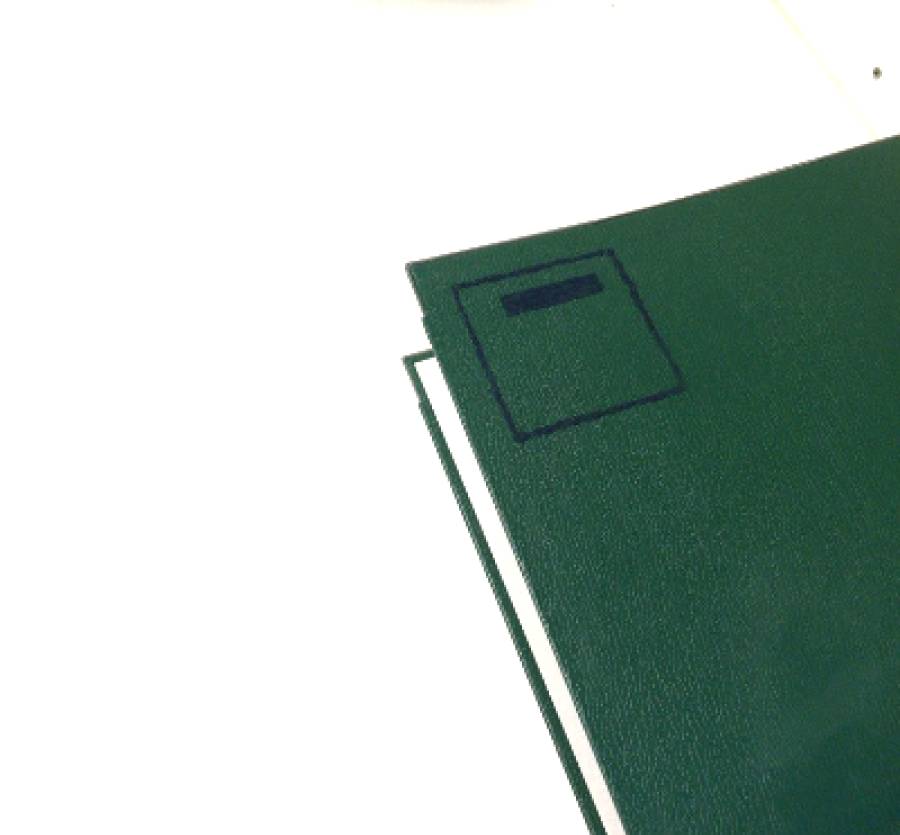}
        \end{subfigure}
        \begin{subfigure}{0.32\columnwidth}
            \includegraphics[width=\linewidth]{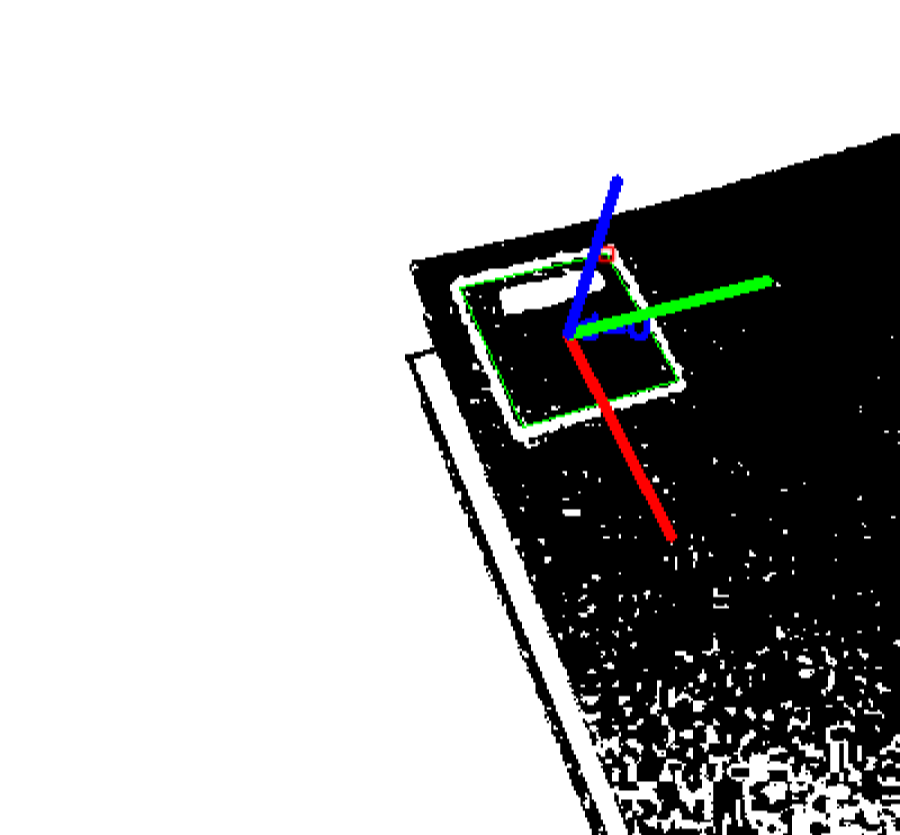}
        \end{subfigure}
        \caption{\scriptsize{Static single-vision setup detecting a visible-range iMarker: raw frame (left), polarizer applied (middle), and detection result using color masking (right).}}
        \label{fig_detection_sv_st}
    \end{subfigure}
    \\[0.8em]
    \begin{subfigure}{0.6\columnwidth}
        \centering
        \begin{subfigure}{0.32\columnwidth}
            \includegraphics[width=\linewidth]{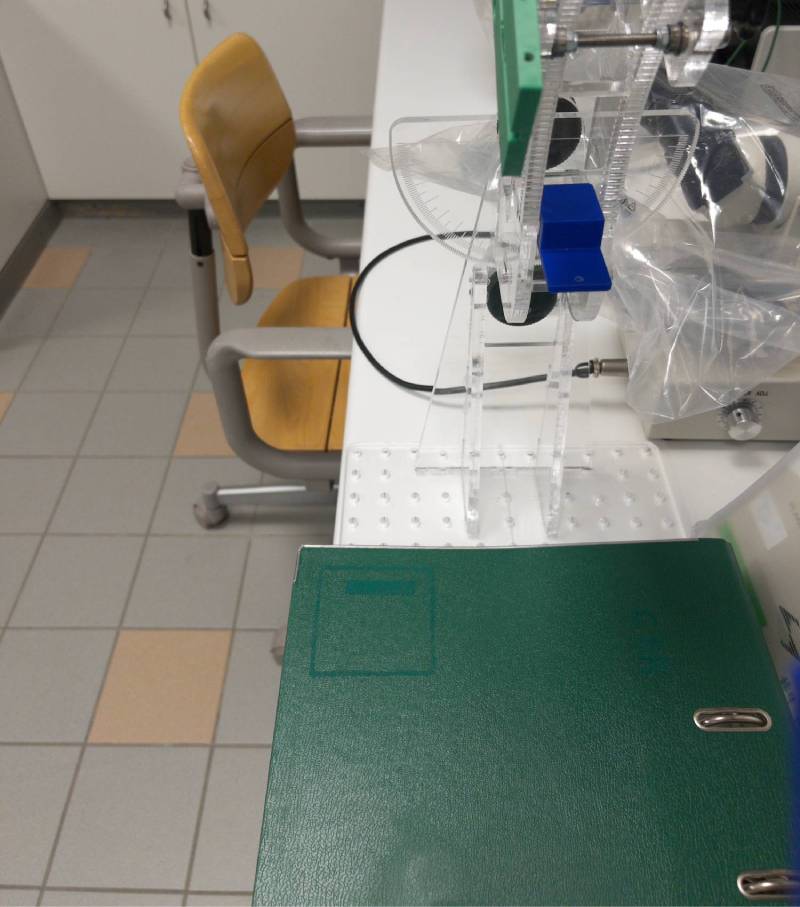}
        \end{subfigure}
        \begin{subfigure}{0.32\columnwidth}
            \includegraphics[width=\linewidth]{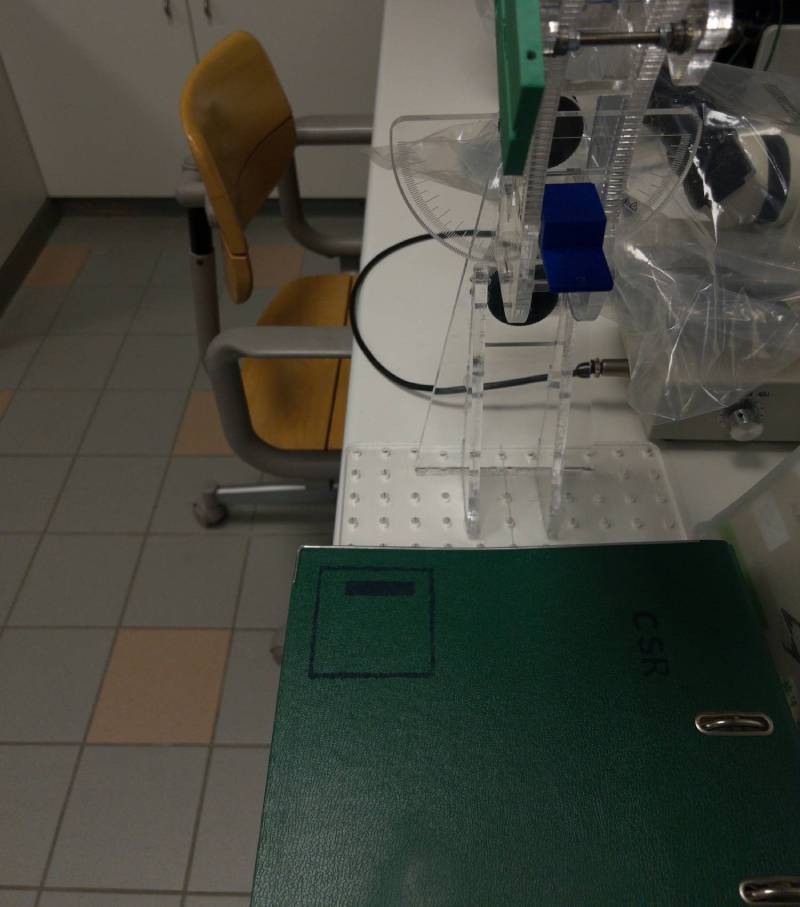}
        \end{subfigure}
        \begin{subfigure}{0.32\columnwidth}
            \includegraphics[width=\linewidth]{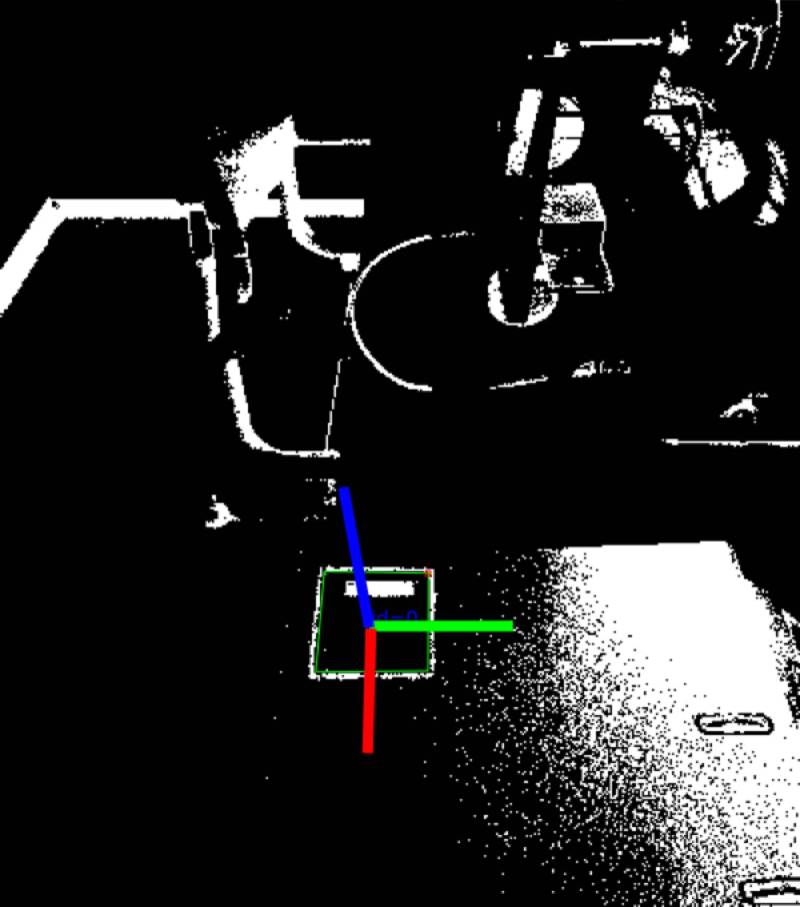}
        \end{subfigure}
        \caption{\scriptsize{Dynamic single-vision setup detecting a visible-range iMarker: frames $f_{t-1}^\text{~off}$ (left) and $f_t^\text{~on}$ (middle) captured at different times with alternating polarizer states, and the detection result obtained via temporal subtraction (right).}}
        \label{fig_detection_sv_dy}
    \end{subfigure}
    \\[0.8em]
    \begin{subfigure}{0.6\columnwidth}
        \centering
        \begin{subfigure}{0.32\columnwidth}
            \includegraphics[width=\linewidth]{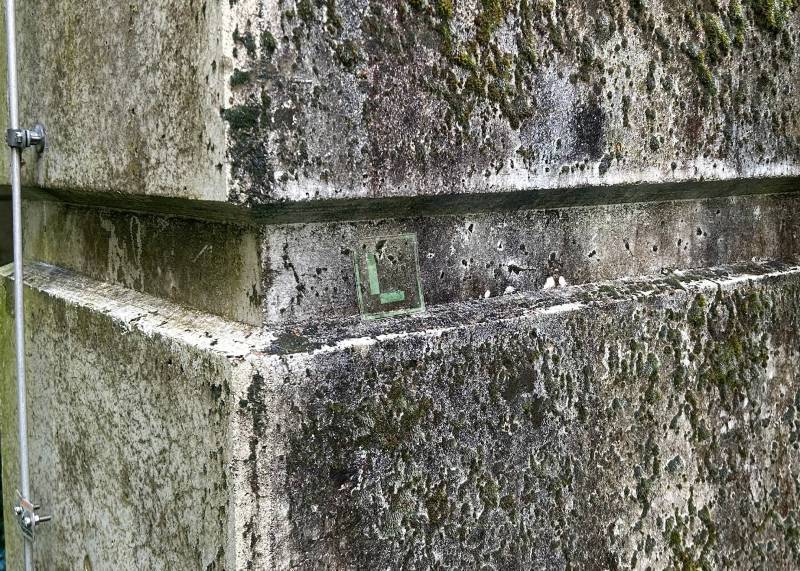}
        \end{subfigure}
        \begin{subfigure}{0.32\columnwidth}
            \includegraphics[width=\linewidth]{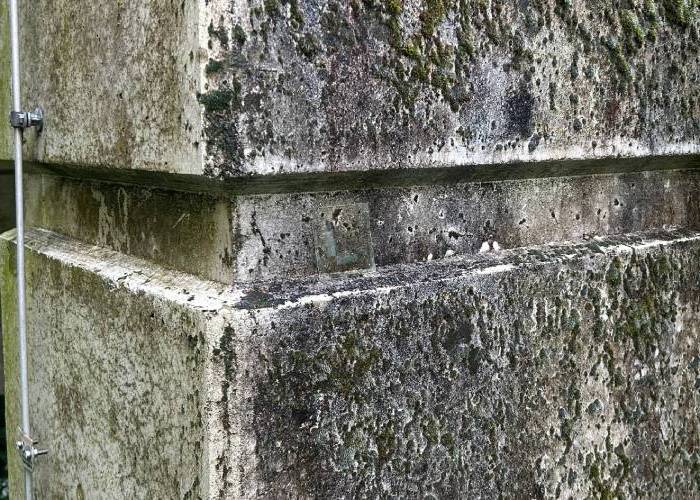}
        \end{subfigure}
        \begin{subfigure}{0.32\columnwidth}
            \includegraphics[width=\linewidth]{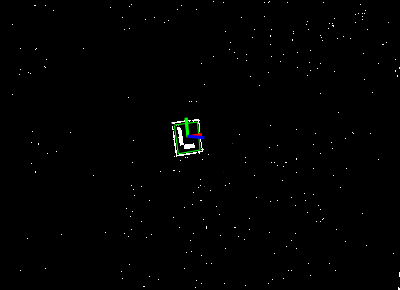}
        \end{subfigure}
        \caption{\scriptsize{Dynamic single-vision setup detecting a visible-range iMarker in an outdoor environment. The order of the frames (off-polarized, on-polarized, and subtraction) follows the same structure as {(b)}.}}
        \label{fig_detection_sv_dy_o}
    \end{subfigure}
    \\[0.8em]
    \begin{subfigure}{0.6\columnwidth}
        \centering
        \begin{subfigure}{0.32\columnwidth}
            \includegraphics[width=\linewidth]{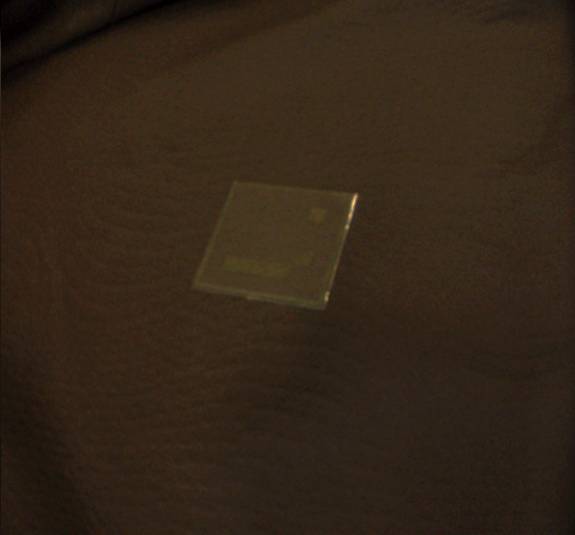}
        \end{subfigure}
        \begin{subfigure}{0.32\columnwidth}
            \includegraphics[width=\linewidth]{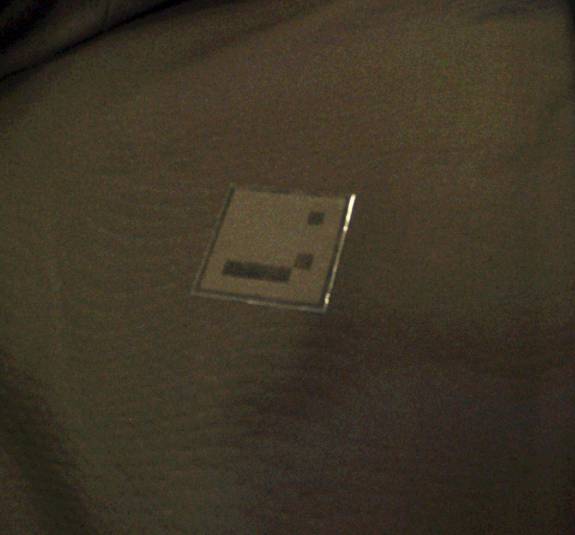}
        \end{subfigure}
        \begin{subfigure}{0.32\columnwidth}
            \includegraphics[width=\linewidth]{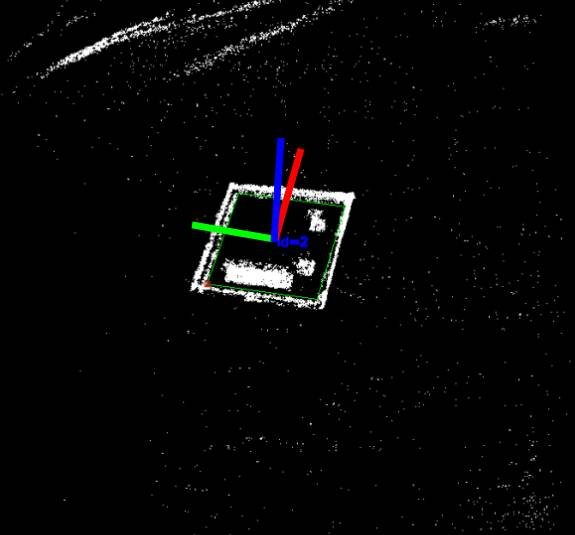}
        \end{subfigure}
        \caption{\scriptsize{Dual-vision setup capturing a visible-range iMarker under low-light conditions: frames $f_l$ (left) and $f_r$ (middle) acquired using orthogonal polarizers, and the detection result obtained via spatial frame subtraction $f_l - f_r$ (right).}}
        \label{fig_detection_dv_pbs}
    \end{subfigure}
    \\[0.8em]
    \begin{subfigure}{0.6\columnwidth}
        \centering
        \begin{subfigure}{0.32\columnwidth}
            \includegraphics[width=\linewidth]{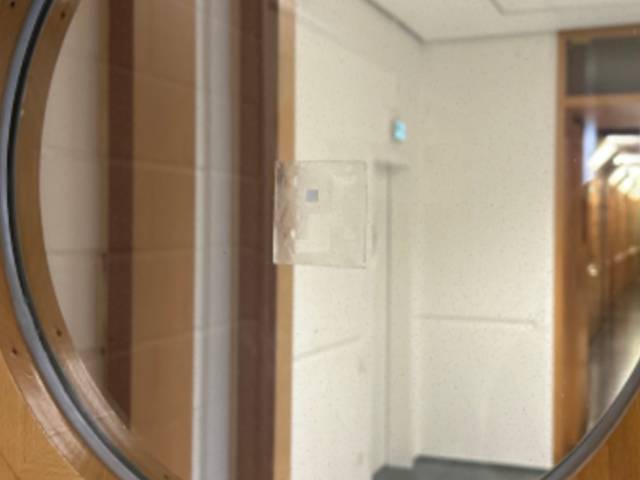}
        \end{subfigure}
        \begin{subfigure}{0.32\columnwidth}
            \includegraphics[width=\linewidth]{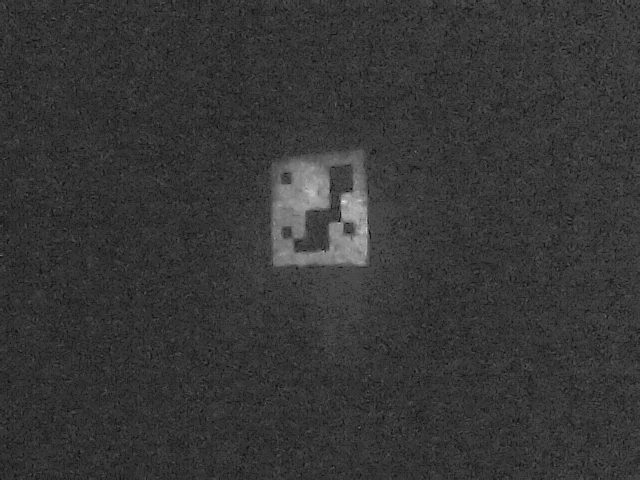}
        \end{subfigure}
        \begin{subfigure}{0.32\columnwidth}
            \includegraphics[width=\linewidth]{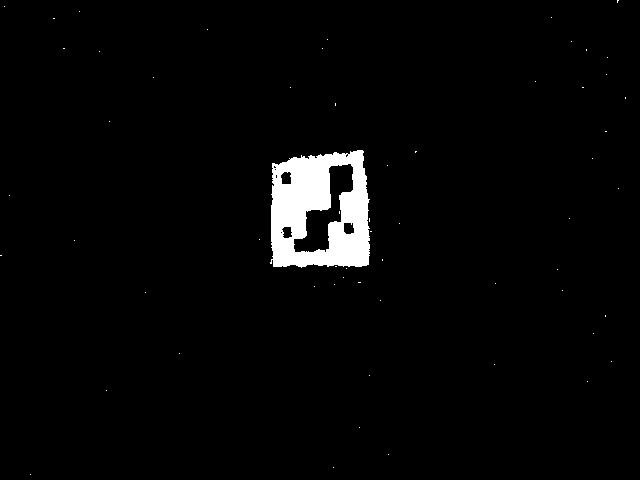}
        \end{subfigure}
        \caption{\scriptsize{Single-vision UV iMarker detection: scene as captured by a standard mobile phone camera (left), corresponding view captured with a UV-sensitive camera under near-UV illumination (middle), and the resulting marker pattern revealing output via thresholding (right).}}
        \label{fig_detection_sv_uv}
    \end{subfigure}
    \caption{iMarker variants detected using different sensor setups and corresponding algorithms, including the visible- and UV-range iMarkers.}
    \label{fig_benchmark}
\end{figure}

%% file: sections/evaluation/detection.tex
\subsection{Detection Range and Pose Accuracy Evaluation}
\label{sec_test_detection}

To evaluate the effectiveness of iMarkers compared with their paper-printed counterparts, a series of comparative experiments was conducted using both marker variants in the ArUco format.
The experiments focused on two primary objectives: (1) determining the maximum detection range across varying viewing angles, and (2) assessing the accuracy of the estimated 6DoF pose for each marker type; both were done using the ArUco detection algorithm.
The employed sensor across all scenarios was the single-vision setup with the RealSense, unmodified for printed markers and equipped with static/dynamic polarizers for iMarker detection.
We intentionally limited the analysis to this setup to ensure a fair comparison for detecting markers, whereas the other sensors in Fig.~\ref{fig_benchmark_hardware} differ significantly in resolution, shutter type, and optical characteristics.
All experiments were conducted using the camera’s RGB sensor at a resolution of $1024 \times 768$ to maintain uniform input quality across evaluations.
Regarding marker dimensions, \(7~\mathrm{cm} \times 7~\mathrm{cm}\) visible-range iMarker variants were used to be compared against \(6~\mathrm{cm} \times 6~\mathrm{cm}\) and \(7~\mathrm{cm} \times 7~\mathrm{cm}\) printed ArUco marker references.
Evaluations were performed under normal lighting conditions in a controlled robotics laboratory equipped with an OptiTrack motion capture system, providing sub-millimeter ground-truth accuracy for 6DoF pose tracking.
Both markers were affixed to a tripod using an adjustable mounting plate, enabling precise control of their orientation relative to the camera.
To ensure accurate ground-truth acquisition during the evaluations, omnireflective motion-capture landmarks were placed on both fiducial markers, and their poses were continuously tracked using the OptiTrack system.
The examined viewing angle ranged from \(0^{\circ}\) to \(75^{\circ}\), within which the estimated poses of the markers remained reliable \textit{w.r.t.} the ground-truth.

\begin{figure}[!t]
    \centering
    \includegraphics[width=.8\columnwidth]{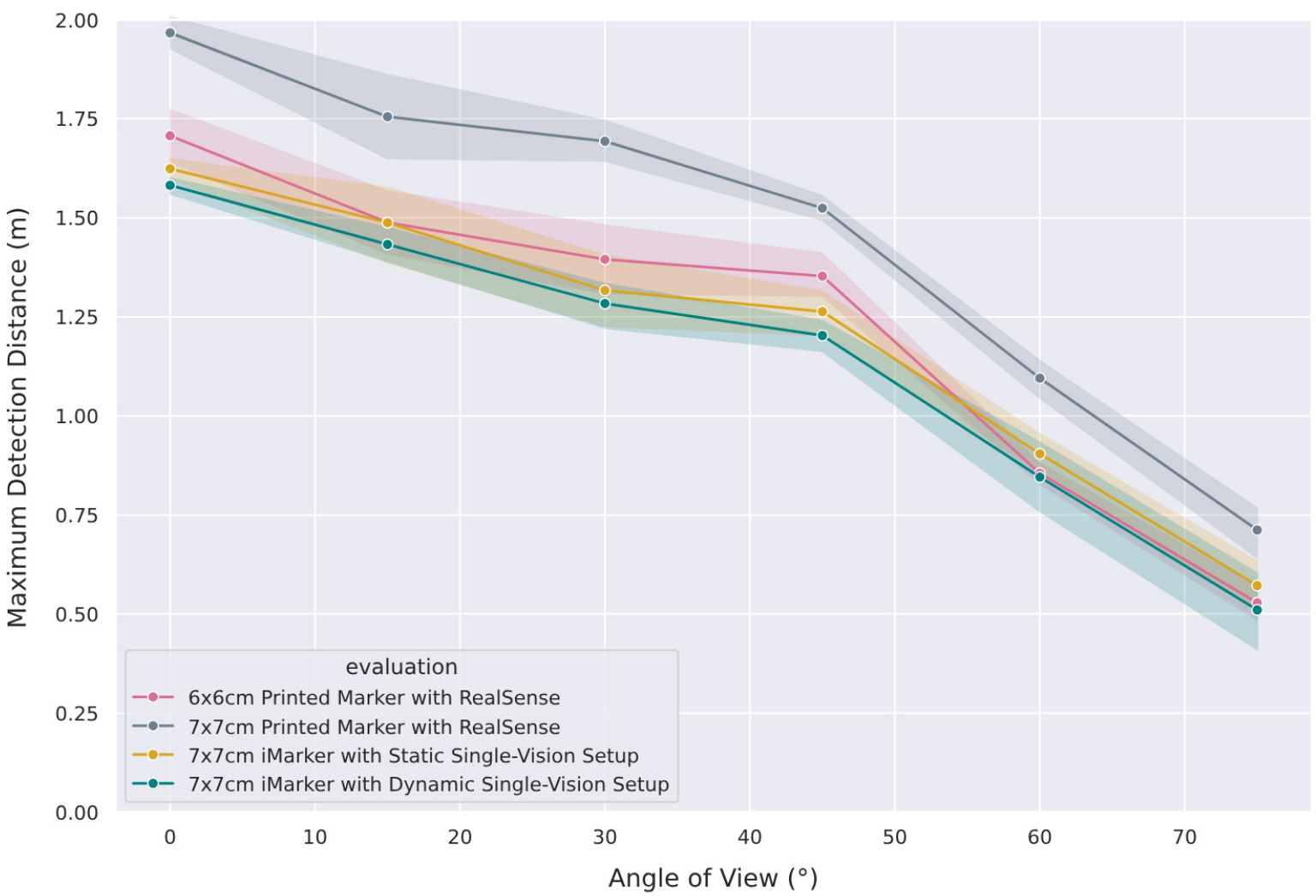}
    \caption{Maximum detection distances for \enquote{printed} and \enquote{iMarkers} ArUco formats at various viewing angles. The shaded regions (error bands) indicate measurement uncertainty in maximum recognition distance.}
    \label{fig_benchmark_detection}
\end{figure}

\noindent\textbf{Maximum detection range.}
According to the evaluation results in Fig.~\ref{fig_benchmark_detection}, increasing the viewing angle reduces the maximum marker detection range in all scenarios, mainly due to perspective distortion.
Additionally, iMarker detection using static and dynamic single-vision setups shows negligible differences in maximum detection ranges, differing between $3~\mathrm{cm}$ to $7~\mathrm{cm}$, with an average of $4.82~\mathrm{cm}$.
Regardless of the detection setup, printed markers consistently achieved higher detection ranges than iMarkers of identical dimensions ($7~\mathrm{cm} \times 7~\mathrm{cm}$).
The detection range of \(7~\mathrm{cm} \times 7~\mathrm{cm}\) iMarkers closely matches that of \(6~\mathrm{cm} \times 6~\mathrm{cm}\) printed ArUco markers, with differences ranging from $4.12~\mathrm{cm}$ to $6.39~\mathrm{cm}$ and averaging $4.24~\mathrm{cm}$ across various angles of view.
It should be noted that printed ArUco markers, thanks to their high-contrast black-and-white patterns, offer substantial visual features that facilitate detection.
However, iMarkers demonstrate comparable detection performance in real-world experiments, apart from their distinct advantage over classic markers: their ability to seamlessly blend into the environment without disrupting visual aesthetics.
This makes iMarkers highly suitable for shorter-range applications where visual discretion is essential, such as indoor navigation.

\input{tables/tbl_pose}

\noindent\textbf{Pose accuracy.}
The pose estimation accuracy was assessed by measuring the deviation between the detected marker poses and the ground-truth at specified viewing angles.
As shown in Table~\ref{tbl_pose_error}, the mean pose estimation error norm across various viewing angles remains consistent among the printed ArUco markers and their iMarker counterparts.
In general, the detection error increases as the viewing angle grows from \(0^{\circ}\) to \(75^{\circ}\)due to increasing perspective distortion of the marker appearance.
The relative difference in pose estimation error between printed markers and iMarkers remains within a narrow range of $1.3\%$ to $3.9\%$.
This consistency and negligible differences in error were expected outcomes of our system design, as the proposed hardware-software configurations are explicitly designed to produce pre-processed binary images that reveal the marker patterns before the \textit{detection and pose estimation} stage in the standard ArUco detector.
Thus, the marker detector library recognizes both markers by applying built-in preprocessing to either standard RGB inputs of printed markers or the binary outputs generated by the iMarker detector.

\noindent\textbf{Results analysis.}
It should be noted that the goal of iMarkers is not to outperform printed markers in all performance metrics, but to provide an imperceptible yet reliably detectable alternative.
Traditional printed markers benefit from high contrast patterns (\textit{e.g.,} black and white) and inherently offer superior detectability than all blended/unobtrusive systems reviewed in Section~\ref{sec_related_blended}.
While the invisibility of iMarkers introduces trade-offs in detection robustness, our experimental evaluations of \textit{maximum detection range} and \textit{pose accuracy} show that iMarkers can still deliver competitive results, while maintaining minimal visual intrusion.

%% file: tables/tbl_pose.tex
\begin{table}[!t]
    \centering
    \caption{Mean pose estimation error (in meters) of \(7 \times 7~\mathrm{cm^2}\) markers across different viewing angles. Results are reported for ``printed markers" and ``iMarkers" (detected using static single-vision (SSV) and dynamic single-vision (DSV) setups) of the ArUco library. Values in brackets indicate the iMarkers' relative increase in error compared to the printed markers at each angle.}
    \begin{tabular}{l|ccc}
        \toprule
            & \multicolumn{3}{c}{\textbf{Marker Variants}} \\
        \midrule
            \textbf{Angle} & \textit{Printed} & \textit{iMarker (SSV)} & \textit{iMarker (DSV)} \\
        \midrule
        \textit{0°}   & 0.155 & 0.160 \tiny{$^{[+3.2\%]}$} & 0.161 \tiny{$^{[+3.9\%]}$} \\
        \textit{15°}  & 0.191 & 0.196 \tiny{$^{[+2.6\%]}$} & 0.194 \tiny{$^{[+1.6\%]}$} \\
        \textit{30°}  & 0.208 & 0.211 \tiny{$^{[+1.4\%]}$} & 0.214 \tiny{$^{[+2.9\%]}$} \\
        \textit{45°}  & 0.214 & 0.218 \tiny{$^{[+1.9\%]}$} & 0.220 \tiny{$^{[+2.8\%]}$} \\
        \textit{60°}  & 0.228 & 0.231 \tiny{$^{[+1.3\%]}$} & 0.234 \tiny{$^{[+2.6\%]}$} \\
        \textit{75°}  & 0.235 & 0.238 \tiny{$^{[+1.3\%]}$} & 0.241 \tiny{$^{[+2.6\%]}$} \\
        \bottomrule
    \end{tabular}
    \label{tbl_pose_error}
\end{table}

%% file: sections/evaluation/low-light.tex
\subsection{Operation in Degraded Visual Conditions}
\label{sec_lowlight}

To further highlight the complementary nature of iMarkers relative to conventional markers, an additional evaluation examines scenarios in which iMarkers remain detectable while printed markers fail.
These experiments target degraded visual conditions, including low-illumination and near-dark environments, where visible-spectrum contrast is insufficient for reliable detection of printed markers.
To assess real-world applicability in robotics, a representative UAV-landing–inspired marker-detection and positioning task was evaluated under varying low-light conditions.
This protocol reflects realistic operational scenarios, such as warehouse automation and night-time inspection, in which robots need to operate reliably under severely degraded visible light.

The comparison is conducted between an IR-range iMarker and a printed ArUco marker of identical physical size ($6~\mathrm{cm} \times 6~\mathrm{cm}$), both placed at a distance of $1~\mathrm{m}$ from the camera.
Markers use the same ArUco dictionary (original) and differ only in their \texttt{marker\_id} to support flexible configurations.
Ambient illuminance was gradually reduced during the experiment by controlling the environmental light sources, with illuminance levels measured using a calibrated lux meter application.
Furthermore, frames captured at matched resolution using the RGB sensor and the IR-range camera with active IR illumination (Ultrafire A100-IR LED torch) and a long-pass filter, with $18$ synchronized frames sampled per illumination level for both RGB and IR streams across all scenarios.
Hence, the RGB stream is used for printed marker detection, and the infrared stream for iMarker detection.

\begin{table}[t]
    \centering
    \caption{Detection robustness of printed ArUco markers ($\mathcal{P}$) and IR-range iMarkers ($\mathcal{I}$) under progressively lower illumination conditions, highlighting scenarios where conventional markers fail but iMarkers remain reliably detectable.}
    \label{tab_low_light_benchmark}
    \begin{tabular}{l | l | c | c | c | c}
        \toprule
            \multicolumn{2}{l|}{} & \multicolumn{4}{c}{\textbf{Illuminance}} \\
        \cmidrule{3-6}
            \multicolumn{2}{l|}{} & $15~\mathrm{lux}$ & $10~\mathrm{lux}$ & $5~\mathrm{lux}$ & $\leq5~\mathrm{lux}$ \\
        \midrule
            \multirow{3}{*}{\shortstack{\textit{Sample} \\ \textit{View}}} & $\mathcal{P}$ & 
            \includegraphics[width=.1\columnwidth]{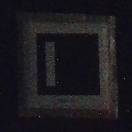} &
            \includegraphics[width=.1\columnwidth]{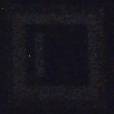} &
            \includegraphics[width=.1\columnwidth]{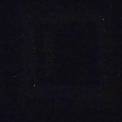} &
            \includegraphics[width=.1\columnwidth]{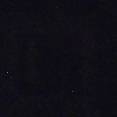} \\
            \cmidrule{2-6}
            & $\mathcal{I}$ & \includegraphics[width=.1\columnwidth]{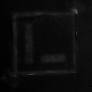} & 
            \includegraphics[width=.1\columnwidth]{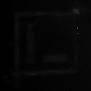} & 
            \includegraphics[width=.1\columnwidth]{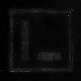} &
            \includegraphics[width=.1\columnwidth]{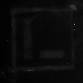} \\
        \midrule
            \multirow{3}{*}{\shortstack{\textit{Sample} \\ \textit{Detection}}} & $\mathcal{P}$ &
            \includegraphics[width=.1\columnwidth]{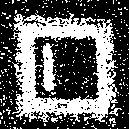} &
            \includegraphics[width=.1\columnwidth]{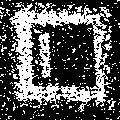} &
            \includegraphics[width=.1\columnwidth]{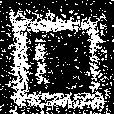} &
            \includegraphics[width=.1\columnwidth]{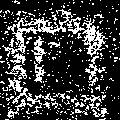} \\
            \cmidrule{2-6}
            & $\mathcal{I}$ & \includegraphics[width=.1\columnwidth]{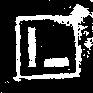} &
            \includegraphics[width=.1\columnwidth]{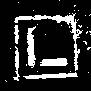} &
            \includegraphics[width=.1\columnwidth]{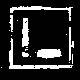} & 
            \includegraphics[width=.1\columnwidth]{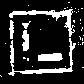} \\
        \midrule
            \multirow{3}{*}{\shortstack{\textit{Detection} \\ \textit{Rate}}} & $\mathcal{P}$ & 0.72 \tiny{(13 / 18)} & 0.50 \tiny{(9 / 18)} & 0.11 \tiny{(2 / 18)} & 0.06 \tiny{(1 / 18)} \\
            \cmidrule{2-6}
            & $\mathcal{I}$ & \textbf{0.89} \tiny{(16 / 18)} & \textbf{0.83} \tiny{(15 / 18)} & \textbf{0.83} \tiny{(15 / 18)} & \textbf{0.78} \tiny{(14 / 18)} \\
        \midrule
            \multirow{3}{*}{\shortstack{\textit{\# False} \\ \textit{Positive}}} & $\mathcal{P}$ & 0.17 \tiny{(3 / 18)} & 0.33 \tiny{(6 / 18)} & 0.33 \tiny{(6 / 18)} & 0.44 \tiny{(8 / 18)} \\
            \cmidrule{2-6}
            & $\mathcal{I}$ & \textbf{0.06} \tiny{(1 / 18)} & \textbf{0.11} \tiny{(2 / 18)} & \textbf{0.17} \tiny{(3 / 18)} & \textbf{0.17} \tiny{(3 / 18)} \\
        \bottomrule
    \end{tabular}
\end{table}

As summarized in Table \ref{tab_low_light_benchmark}, various complementary metrics, including detection rate and false positive frequency, have been analyzed for printed markers and IR-range iMarkers under progressively decreasing illumination conditions.
Here, the detection rate is defined as the ratio of frames in which the correct \texttt{marker\_id} is detected to the total number of frames captured, all at illumination level $l$.
False positives are incorrect detections triggered by high-contrast background textures that resemble marker patterns, such as repetitive grids or net-like structures.

According to the table, the \textit{detection rate} of printed markers degrades sharply as illumination decreases, because reduced contrast causes marker features to blend into background noise.
Specifically, performance drops from $0.72$ at $15~lux$ to 0.06 at $\leq5~\mathrm{lux}$.
In contrast, IR iMarkers maintain a largely stable detection rate, varying only slightly from 0.89 to 0.78, with minor degradation attributed to sensor noise and IR signal attenuation under extremely low-light conditions.
A similar trend is observed for \textit{false positives}: as illumination decreases, erroneous detections for printed markers increase substantially (from 0.17 to 0.44), primarily due to the reduced \textit{signal-to-noise ratio} in the RGB stream.
In contrast, iMarkers maintain a consistently low false-positive rate across all illumination levels, with only minor fluctuations, owing to the CSRs' distinctive IR reflectance properties.

These results show that while printed markers are strongly affected by illumination-dependent failure modes, iMarkers provide reliable detection performance under degraded visual conditions.
Additionally, it clearly demonstrates that iMarkers go beyond \textit{aesthetic considerations} and act as an \textit{enabling technology} for robotic perception in scenarios where conventional fiducial markers become unreliable or unusable.

%% file: sections/evaluation/recog_speed.tex
\subsection{Recognition Speed}
\label{sec_speed}

\input{tables/tbl_profiling}

Given the need for fast recognition to demonstrate the practicality of fiducial markers for real-time applications (in our context, with an average processing rate of $25 \pm 5 \mathrm{fps}$), this subsection presents a detailed profiling analysis of iMarker detection and recognition.
Table~\ref{tbl_profiling} summarizes cumulative processing times for iMarker data decoding.
Evaluations were conducted using various sensor configurations and implemented software introduced in subsection~\ref{sec_implement}.
The algorithms developed for all sensor variants share five core modules: 
\enquote{\textit{acquisition},} which captures video frames;
\enquote{\textit{preprocessing},} which applies image enhancement techniques;
\enquote{\textit{processing},} which refines frames to emphasize marker patterns;
\enquote{\textit{post-processing},} which enhances the final binary images; and
\enquote{\textit{recognition}} to detect and interpret marker patterns.
To obtain the measurements, \(7~\mathrm{cm} \times 7~\mathrm{cm}\) iMarkers were placed approximately $50~\mathrm{cm}$ from the lens, directly facing the sensor under standard lighting conditions with evenly diffused artificial illumination.
For the \ac{UV}-range iMarker, a DARKBEAM A300 \(365~\mathrm{nm}\) flashlight was additionally used to enhance the iMarker visibility.

According to Table~\ref{tbl_profiling}, the overall iMarker recognition process, from video frame \enquote{\textit{acquisition}} to inner pattern \enquote{\textit{recognition}} steps, is completed within milliseconds and adds only $\approx 100~ms$ overhead to the ArUco library.
The reported results are obtained using standard CPU-based processing and without the aid of GPU or other hardware accelerators.
For comparison, some methods discussed in Table~\ref{tbl_markers} require several seconds for detection, underscoring the efficiency of the iMarker system.
Among the various configurations, \ac{UV}-range iMarker recognition (Algorithm~\ref{alg_single_thresh}) is the fastest, as the input image is already in grayscale, requiring only a simple \enquote{thresholding} operation to reveal the marker pattern for recognition.
In dual-vision setups (Algorithm~\ref{alg_double_subtract}), \enquote{\textit{acquisition}} and \enquote{\textit{processing}} account for the subsequent significant portions of the processing time due to the need to receive synchronized frames from two cameras and perform image alignment for further processing, respectively.
The \enquote{\textit{post-processing}} stage is executed more efficiently because the substantial filtering process retains only the regions that differ, \textit{i.e.,} the \ac{CSR} areas.
On the other hand, single-vision setups are faster during \enquote{\textit{acquisition}} because they capture frames from a single camera.
However, they allocate more time to \enquote{\textit{post-processing}}, as the captured images contain higher noise levels and require effective refinement of iMarker regions.

%% file: tables/tbl_profiling.tex
\begin{table*}[!t]
    \centering
    \caption{Performance analysis of iMarkers detection and recognition, based on cumulative time profiling of required modules (averaged over 40 frames) in milliseconds $(\mathrm{ms})$. Abbreviations are detailed below the table.}
    \begin{tabular}{l|c|c|c|c|c}
        \toprule
            & \multicolumn{5}{c}{\textbf{iMarker Variant}} \\
        \cmidrule{2-6}
            & \textit{UV} & \multicolumn{4}{c}{\textit{Visible-range}} \\
        \midrule
            \textbf{Software Module} & \textbf{SV Static} & \textbf{SV Static} & \textbf{SV Dynamic} & \textbf{DV (cube BS)} & \textbf{DV (plate BS)} \\
        \midrule
            \textit{Acquisition}      & \cellcolor{greenl}{\(11~\mathrm{ms}\)} & \cellcolor{greenl}{\(15~\mathrm{ms}\)} & \cellcolor{greenl}{\(13~\mathrm{ms}\)} & \cellcolor{yellow}{\(32~\mathrm{ms}\)} & \cellcolor{yellow}{\(36~\mathrm{ms}\)} \\
            \textit{Preprocessing}    & \cellcolor{greend}{\(2~\mathrm{ms}\)} & \cellcolor{greend}{\(7~\mathrm{ms}\)} & \cellcolor{greend}{\(9~\mathrm{ms}\)} & \cellcolor{greenl}{\(14~\mathrm{ms}\)} & \cellcolor{greenl}{\(19~\mathrm{ms}\)} \\
            \textit{Processing}       & \cellcolor{greend}{\(3~\mathrm{ms}\)} & \cellcolor{greenl}{\(23~\mathrm{ms}\)} & \cellcolor{greenl}{\(29~\mathrm{ms}\)} & \cellcolor{yellow}{\(39~\mathrm{ms}\)} & \cellcolor{yellow}{\(42~\mathrm{ms}\)} \\
            \textit{Post-processing}  & \cellcolor{greend}{\(3~\mathrm{ms}\)} & \cellcolor{greend}{\(9~\mathrm{ms}\)} & \cellcolor{greend}{\(8~\mathrm{ms}\)} & \cellcolor{greend}{\(4~\mathrm{ms}\)} & \cellcolor{greend}{\(5~\mathrm{ms}\)} \\
            \textit{Recognition}      & \cellcolor{greend}{\(2~\mathrm{ms}\)} & \cellcolor{greend}{\(2~\mathrm{ms}\)} & \cellcolor{greend}{\(3~\mathrm{ms}\)} & \cellcolor{greend}{\(2~\mathrm{ms}\)} & \cellcolor{greend}{\(3~\mathrm{ms}\)} \\
        \midrule
            \textbf{Sum}     & \(21~\mathrm{ms}\) & \(56~\mathrm{ms}\) & \(62~\mathrm{ms}\) & \(91~\mathrm{ms}\) & \(105~\mathrm{ms}\) \\
        \bottomrule
        \multicolumn{5}{l}{$^{\mathrm{*}}$PS. \textit{SV}: Single-Vision, \textit{DV}: Dual-Vision, \textit{UV}: Ultraviolet, \textit{BS}: Beamsplitter}
    \end{tabular}
    \label{tbl_profiling}
\end{table*}

%% file: sections/05-discussion.tex
\section{Discussions}
\label{sec_discuss}


\subsection{Comparative Insights}
\begin{figure}[t]
     \centering
     \includegraphics[width=.8\columnwidth]{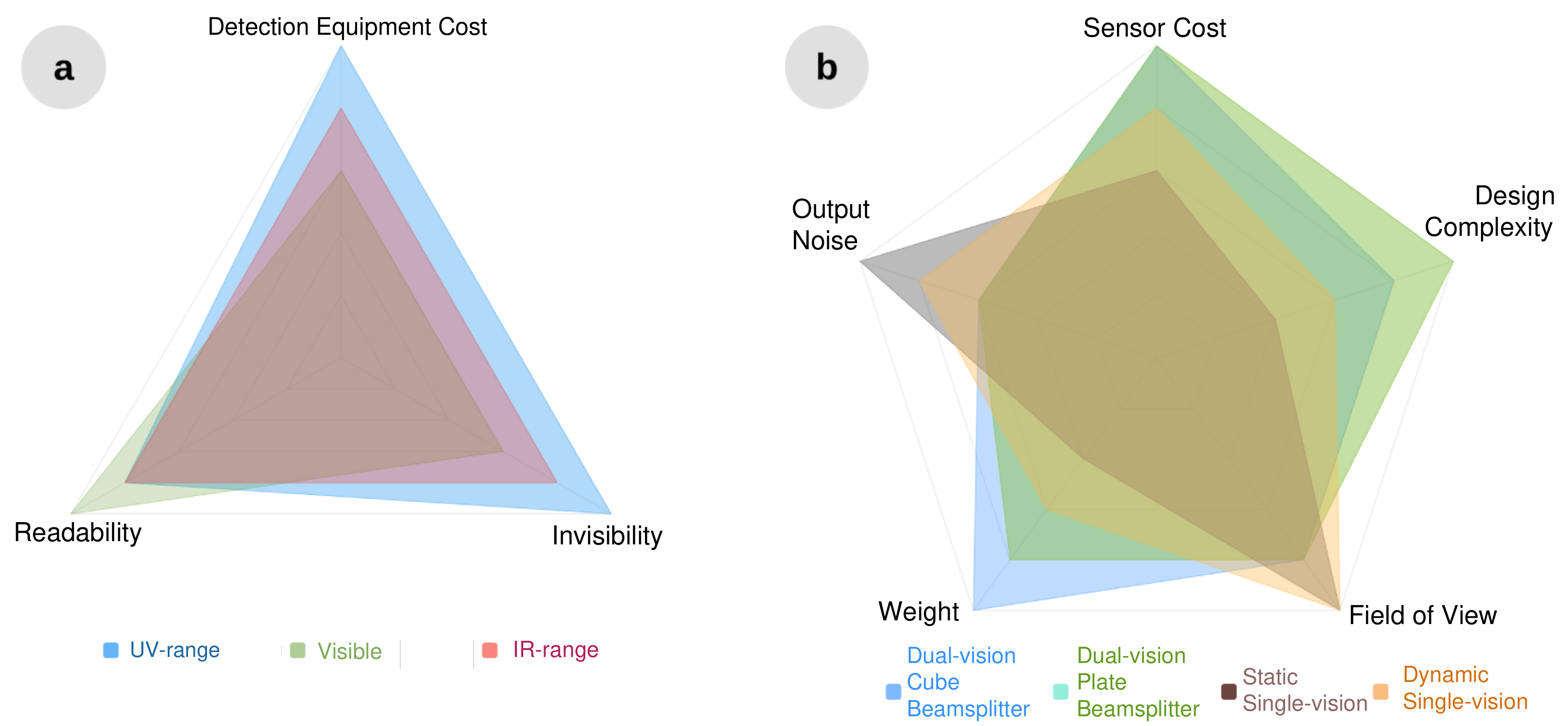}
     \caption{Relative performance scores of iMarker variants (\textbf{a}) and sensors (\textbf{b}) across multiple aspects, with larger areas indicating higher values. The term \enquote{BS} stands for beamsplitter.}
     \label{fig_sensor_matrix}
\end{figure}

Fig.~\ref{fig_sensor_matrix} depicts relative performance scores of different iMarker fabrication variants and the designed sensor configurations for their detection, where larger areas indicate higher values.
The relative performance scores were assigned based on theoretical analysis and experimental observations, which will be described in detail.

In Fig.~\ref{fig_sensor_matrix}a, three primary performance aspects of the iMarker variants have been analyzed.
Regarding \enquote{invisibility,} \ac{UV}-range iMarkers remain entirely imperceptible to the eye, while \ac{IR}-range iMarkers may exhibit slight visibility due to scattering under extreme lighting conditions.
Visible-range iMarkers, on the other hand, are designed to blend into patterned backgrounds; however, upon close inspection, they remain minimally visible to the human eye.
When \enquote{readability} is studied, visible-range iMarkers have better compatibility with standard cameras and do not require additional illumination.
However, \ac{IR}- and \ac{UV}-range iMarkers require dedicated illumination sources and specific sensors, and their detection performance varies with environmental conditions.
Finally, and in terms of \enquote{detection equipment cost} (camera, flashlight, \textit{etc.}), visible-range iMarkers are the most economical, as they can be detected using standard cameras.
\ac{UV} equipment (camera and illumination sources) is generally the most expensive of the three, and \ac{IR}-range iMarker detection equipment costs lie between them.

Considering the sensor-related factors shown in Fig.~\ref{fig_sensor_matrix}b, the performance scores can be studied in five aspects.
In \enquote{design complexity,} the static single-vision setup is the most straightforward, requiring only a static polarizer attached to the lens.
The dynamic single-vision setup has moderate complexity, incorporating a function generator to control polarization.
Dual-vision configurations are more complex, with the cube beamsplitter offering more reliable light division than the plate beamsplitter, which requires precise alignment for optimal splitting.
The design complexity directly influences the \enquote{sensor cost.}
The static single-vision setup is the most economical, requiring only a polarizer, whereas the dynamic single-vision configuration incurs additional costs due to its polarization-control components.
Dual-vision setups are generally more expensive: they require two cameras, a beamsplitter (either cube or plate), and a structural mounting cage to ensure precise alignment and stability.

Regarding the \enquote{field of view,} single-vision setups provide a broader theoretical \ac{FoV} (\textit{e.g.,} \(87^{\circ} \times 58^{\circ}\) in the RealSense™ D435, shown in Fig.~\ref{fig_benchmark_hardware}c).
However, in dual-vision setups, the beamsplitter's physical dimensions and the supporting cage partially obstruct the cameras' \ac{FoV}, reducing the effective viewing area.
For instance, the configuration shown in Fig.~\ref{fig_benchmark_hardware}-c achieves an effective \ac{FoV} of \(57^{\circ} \times 54.5^{\circ}\).
Another essential factor is \enquote{weight}; the static single-vision setup is the lightest, whereas the dynamic single-vision setup adds approximately \(30~\mathrm{gr}\) due to the function generator and dynamic polarizer.
Thus, single-vision setups are more feasible for minimizing payload weight, \textit{e.g.,} in drones, as dual-vision configurations are heavier.
The combined weight of dual-vision setup components typically ranges from \(150\) to \(350~\mathrm{gr}\), with the plate beamsplitter potentially reducing the overall sensor weight.

When choosing among sensor configurations, a key selection criterion is the trade-off between ``detection robustness" and ``clutter (noise) rejection," as each configuration’s detection pipeline can be tailored to optimize performance for specific application scenarios. Dual-vision setups rely on spatial frame subtraction followed by thresholding, producing cleaner outputs by isolating CSR-coated iMarker regions while effectively suppressing non-marker backgrounds thanks to the differential visibility of CSRs between the two sensor views. However, they are more sensitive to artifacts caused by sensor noise, motion, illumination variations, and frame misalignment. In contrast, single-vision setups may retain more background clutter (noise pixels) alongside detected iMarkers. Accordingly, enhancing robustness and output quality is addressed differently depending on the sensor variant. While strategies such as adaptive thresholding and active frame registration can better aid dual-vision setup algorithms, single-vision systems can benefit from lightweight morphological filtering to enhance detection quality without compromising system simplicity.

\subsection{Potential Applications}
Considering the unique characteristics of iMarkers, they serve as potential alternatives in scenarios reliant on printed markers.
In this regard, the obscure nature of iMarkers makes them appropriate for surroundings with unwanted visual clutter, such as museums, hospitals, offices, \textit{etc.}, and for scenarios involving transparent or reflective structures (\textit{e.g.,} glass walls and partitions), where placing printed fiducial markers is impractical or unacceptable due to their high visibility and disruption of visual transparency.
By safely replacing printed markers, iMarkers can preserve unobtrusiveness and aesthetic integrity while delivering full functionality for robotics and \ac{AR} systems.
Note that various AR applications, particularly those deployed on handheld mobile devices, rely exclusively on RGB cameras and often use fiducial landmarks to support interactive features such as user engagement.
This aligns well with the iMarker detection procedure proposed in this paper; specifically, the single-vision setup, which requires minimal hardware modifications to enable compatibility with standard AR-capable vision sensors.

Additionally, as the iMarker detection pipeline relies on simple computer vision techniques (\textit{e.g.,} thresholding and frame subtraction), it introduces minimal computational overhead and operates efficiently even on a CPU-based system, without requiring hardware acceleration.
Thus, the core innovation of the proposed system lies in tightly coupling hardware and software: the sensor hardware captures frames with filtered CSR-coated iMarker regions, and the software extracts and reveals these regions before passing the results to standard fiducial marker libraries (\textit{e.g.,} ArUco or AprilTag).

\begin{figure*}[!t]
    \centering
    \begin{subfigure}{0.48\textwidth}
        \centering
        \includegraphics[width=\columnwidth]{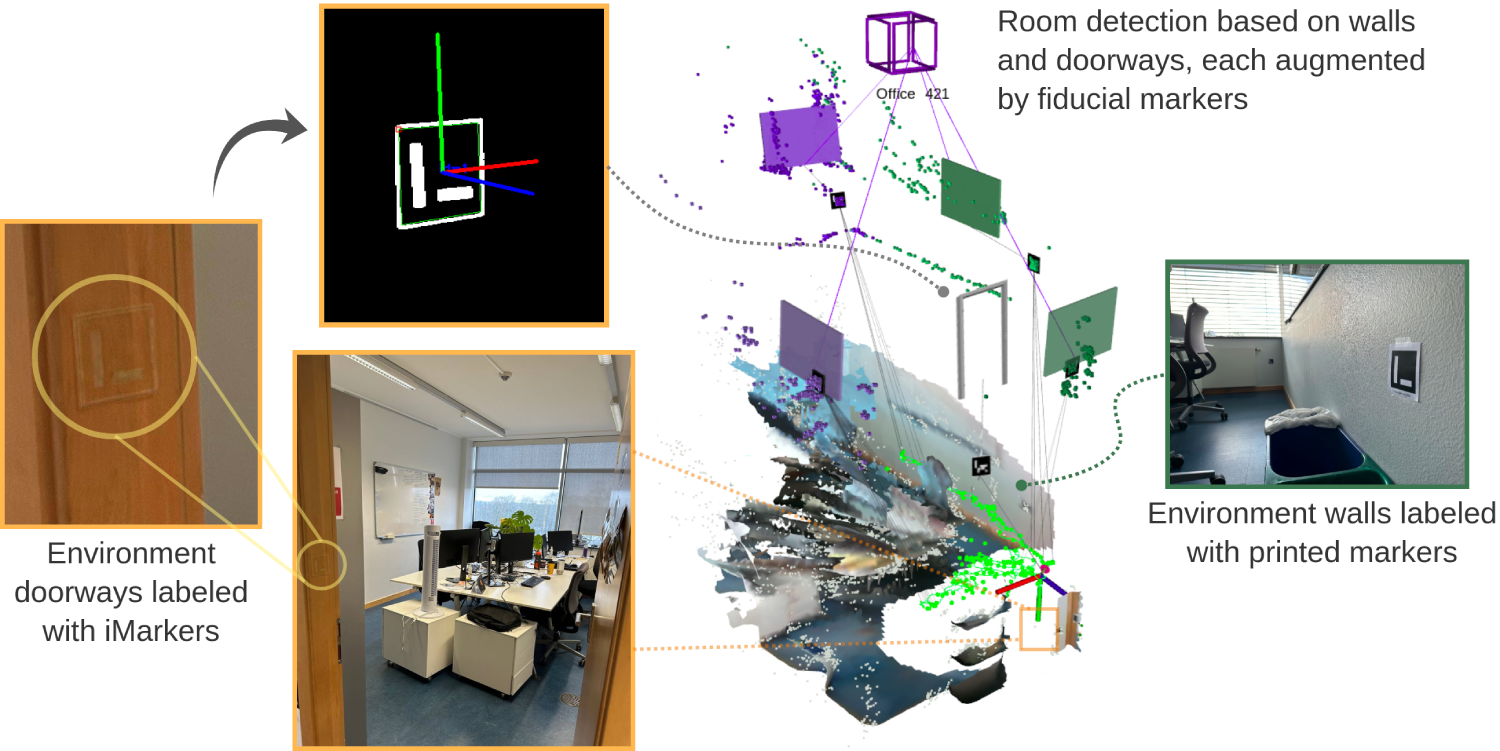}
        \caption{\scriptsize{iMarkers integrated into a visual SLAM pipeline to support camera localization and semantic annotation of indoor environments \cite{vsgraphs1}.}}
        \label{fig_benchmark_robotics_vslam}
    \end{subfigure}
    \hfill
    \begin{subfigure}{0.48\textwidth}
        \centering
        \includegraphics[width=\columnwidth]{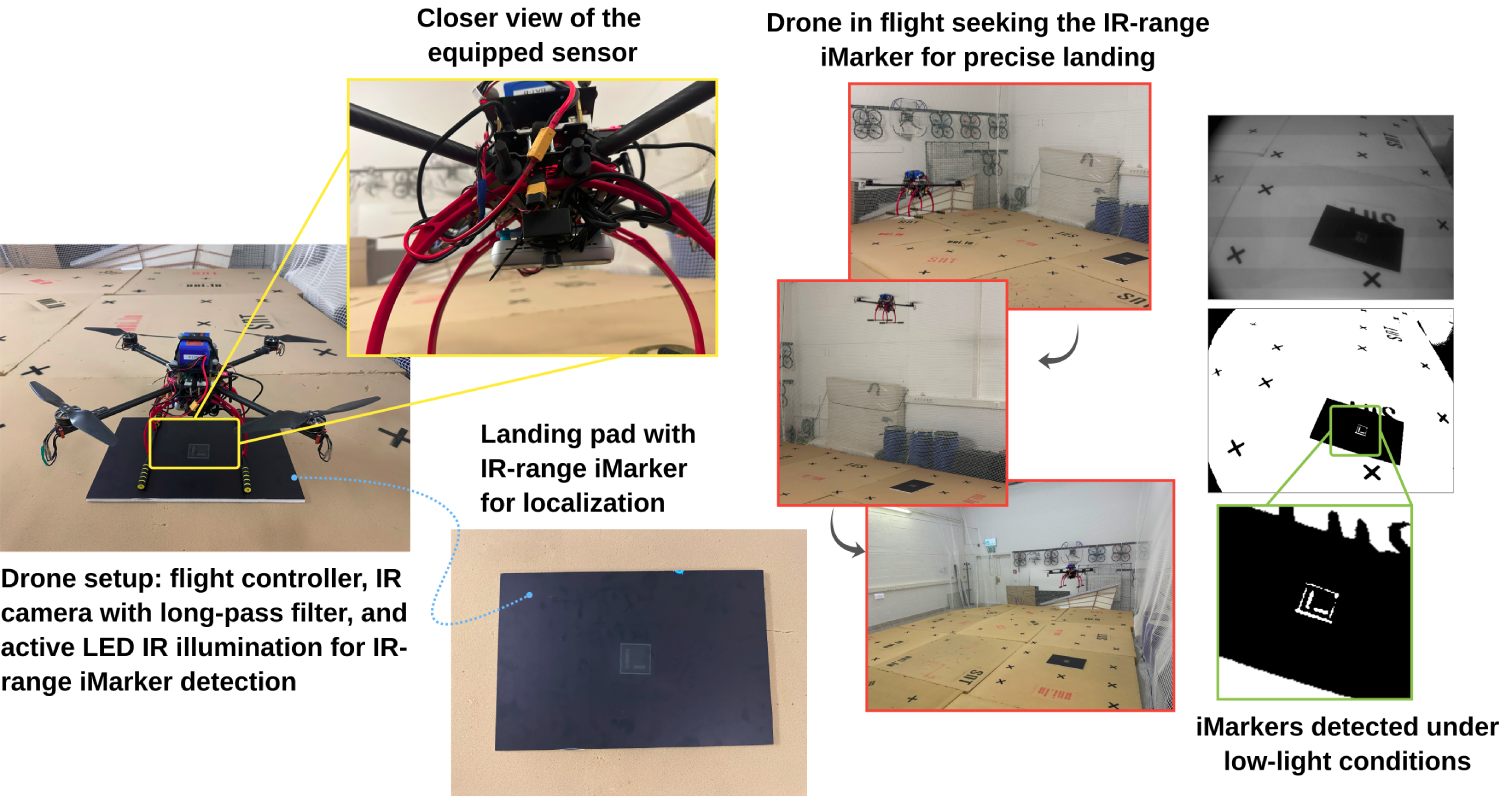}
        \caption{\scriptsize{iMarkers deployed as low-illumination landing targets, enabling reliable pose estimation and precise autonomous landing of aerial robots.}}
        \label{fig_benchmark_robotics_land}
    \end{subfigure}
    \caption{Real-world robotics applications enabled by iMarkers.}
    \label{fig_benchmark_robotics}
\end{figure*}

Fig.~\ref{fig_benchmark_robotics} depicts two representative real-world robotics applications in which iMarkers play a crucial role.
In Fig.~\ref{fig_benchmark_robotics_vslam}, iMarkers are benchmarked within a \ac{VSLAM} framework, showcasing the use of visible-range landmarks to support indoor map reconstruction enriched with semantic information \cite{vsgraphs1}.
Accordingly, iMarkers are applied to office door frames to unobtrusively encode semantic cues related to structural entities, thereby enabling semantic mapping while remaining visually camouflaged within the environment.
In another setting, Fig.~\ref{fig_benchmark_robotics_land} shows a UAV-based robot-landing scenario introduced in Section~\ref{sec_lowlight}, where conventional printed markers become unreliable or undetectable under low-illumination conditions, while IR-range iMarkers remain robustly detectable, enabling reliable localization and safe landing despite severely degraded visible-light cues.

\subsection{Limitations and Fabrication Considerations}
Given the ongoing nature of this research and the absence of a rapid iMarker fabrication process to date, the experiments remain constrained by varying illumination conditions, weather (for outdoor applications), and environmental factors (\textit{e.g.,} dirt).
In terms of environmental robustness, iMarkers are expected to be susceptible to dirt and occlusion, as with conventional printed markers, since both rely on visual input for detection and pose estimation.
In other words, the marker recognition pipeline (ArUco in this work) processes either RGB frames from standard cameras or the binary outputs of the iMarker detector, without introducing additional sensitivity.
Regarding sunlight and outdoor use, UV-range iMarkers may experience reduced contrast under ambient sunlight, potentially leading to false positives due to uncontrolled UV reflections.
Accordingly, the reflectivity of UV-range CSRs needs to be enhanced to ensure that all UV-range iMarkers are reliably detected when illuminated by the light source.
Similarly, IR-range iMarkers could be influenced by environmental IR sources.
In real-world deployments, additional filtering strategies or controlled illumination may be required to ensure reliable iMarker detection.

The scalability of iMarker fabrication for mass deployment remains an active area of investigation.
Currently, production involves several manual steps, including microfluidic droplet generation, annealing, polymerization, and sequential washing, which reduce throughput and introduce variability in marker quality.
To overcome this, we are developing an automated production platform that combines all stages into a continuous process with real-time computer monitoring.
This system is expected to streamline manufacturing, improve consistency, and enable large-scale high-throughput fabrication suitable for real-world robotics applications.

%% file: sections/06-conclusion.tex
\section{Conclusions}
\label{sec_conclusions}

This paper introduced a novel generation of fiducial markers, termed \textit{iMarkers}, designed explicitly for real-world robotics applications.
iMarkers are designed to be visually indistinguishable to the human eye, ensuring minimal environmental disruption while remaining fully detectable and recognizable by robots equipped with appropriate sensors and detection algorithms.
The paper also presents various sensor designs and implemented algorithms for detecting iMarkers placed in the environment and exploiting them for different use cases.
Experimental results demonstrate the practicality and potential of these markers as versatile landmarks for data embedding in various real-world scenarios.

Future work includes optimizing the iMarker fabrication process to ensure scalable production and enhancing illumination equipment, sensor designs, and algorithm implementations to ensure robustness across various scenarios.